\documentclass[preprint,3p,onecolumn]{elsarticle}

\usepackage{graphicx}

\usepackage{color,xspace}
\usepackage{amsmath,amssymb}
\usepackage{multirow}
\usepackage{subfigure, enumerate}
\usepackage[ruled,lined,linesnumbered,nofillcomment]{algorithm2e}

\providecommand{\abs}[1]{\lvert#1\rvert} 
\DeclareMathOperator*{\med}{med}
\providecommand{\abs}[1]{\lvert#1\rvert}

\newcommand{\sammy}   {{\sc Samantha}\xspace }
\newcommand{\sam}     {{struc\-ture-and-motion}\xspace }
\newcommand{\vsfm}    {{\sc VisualSFM}\xspace }

\journal{Computer Vision and Image Understanding}

\begin{document}

\begin{frontmatter}

\title{Hierarchical structure-and-motion recovery from uncalibrated images}

\author{Roberto Toldo\fnref{fn1}}
\ead{roberto.toldo@3dflow.net}
\fntext[fn1]{R.T. is now with 3Dflow s.r.l.~Verona, Italy}

\author{Riccardo Gherardi\fnref{fn2}}
\ead{riccardo.gherardi@crl.toshiba.co.uk}
\fntext[fn2]{R.G. is now with Toshiba Cambridge Research Laboratory}

\author{Michela Farenzena\fnref{fn3}}
\address{Dipartimento di Informatica, Universit\`a di Verona, \\
	Strada Le Grazie, 15 - 37134 Verona, Italy}
\ead{michela.farenzena@evsys.net}
\fntext[fn3]{M.F. is now at  is now with eVS s.r.l.}

\author{Andrea Fusiello\corref{mycorrespondingauthor}}
\cortext[mycorrespondingauthor]{Corresponding author}
\address{
Dipartimento di Ingegneria Elettrica, Gestionale e Meccanica,\\
University of Udine, Via Delle Scienze, 208 - 33100 Udine, Italy\\
}
\ead{andrea.fusiello@uniud.it}

\begin{abstract}
This paper addresses the \sam problem, that requires to find camera motion and
3D structure from point matches. A new pipeline, dubbed \sammy, is presented, that
departs from the prevailing sequential paradigm and embraces instead a
hierarchical approach. This method has several advantages, like a provably lower
computational complexity, which is necessary to achieve true scalability, and
better error containment, leading to more stability and less drift.
Moreover, a practical autocalibration procedure allows to process images without
ancillary information. Experiments with real data assess the accuracy and the
computational efficiency of the method.
\end{abstract}


\end{frontmatter}

\section{Introduction}

The progress in three-dimensional (3D) modeling research has been rapid and
hectic, fueled by recent breakthroughs in keypoint detection and matching, the
advances in computational power of desktop and mobile devices, the advent of
digital photography and the subsequent availability of large datasets of public
images.  Today, the goal of definitively bridging the gap between physical reality
and the digital world seems within reach given the magnitude, breadth and scope
of current 3D modeling systems.

Three dimensional modeling is the process of recovering the properties of the
environment and optionally of the sensing instrument from a series of measures.
This generic definition is wide enough to accommodate very diverse
methodologies, such as time-of-flight laser scanning, photometric stereo or
satellite triangulation.  The \sam (a.k.a.~structure-from-motion) field of
research is concerned with the recovery of the three dimensional geometry of the
scene (the structure) when observed through a moving camera (the motion). Sensor
data is either a video or a set of exposures; additional informations, such as
the calibration parameters, can be used if available.  This paper describes
our contributions to the problem of \sam recovery from unordered, uncalibrated
images i.e, the problem of building a three dimensional model of a scene given a
set of exposures. The sought result (the ``model'') is generally a 3D
point-cloud consisting of the points whose projection was identified and
matched in the images and a set of camera matrices, identifying position and
attitude of each image with respect to an arbitrary reference frame.

The main challenges to be solved are computational efficiency (in order to be
able to deal with more and more images) and generality, i.e., the amount of
ancillary information that is required. 

To address the efficiency issue we propose to describe the entire \sam process as
a binary tree  (called \emph{dendrogram})  constructed by
agglomerative clustering over the set of images. Each leaf corresponds to a
single image, while internal nodes represent partial models obtained by merging the
left and right sub-nodes. Computation proceeds from bottom to top, starting from
several seed couples and eventually reaching the root node, corresponding to the
complete model.  This scheme provably cuts the computational
complexity by one order of magnitude (provided that the dendrogram is well
balanced), and it is less sensitive to typical problems of sequential approaches,
namely sensitivity to initialization \cite{ThoBroWei04} and drift
\cite{CorEA08}. It is also scalable and efficient, since it partitions the
problem into smaller instances and combines them hierarchically, making it
inherently parallelizable.

On the side of generality, our aim is to push the ``pure'' \sam technique as far
as possible, to investigate what can be achieved without including any auxiliary
information.  Current \sam research has partly sidestepped this issue using
ancillary data such as EXIF tags embedded in conventional image formats. Their
presence or consistency, however, is not guaranteed.  We describe our
approach to autocalibration, which is the process of automatic estimation from
images of the internal parameters of the cameras that captured them, and we 
therefore demonstrate the first \sam pipeline capable of using unordered,
uncalibrated images.

\subsection{Structure-and-motion: related work.}

The main issue to address in \sam is the computational complexity, which is
dominated by the bundle adjustment phase, followed by feature extraction and
matching.

A class of solutions that have been proposed are the so-called partitioning
methods \cite{FitZis98}. They reduce the \sam problem into smaller and better
conditioned subproblems which can be effectively optimized. Within this approach,
two main strategies can be distinguished.

The first one is to tackle directly the bundle adjustment algorithm, exploiting
its properties and regularities. The idea is to split the optimization problem
into smaller, more tractable components. The subproblems can be selected
analytically as in \cite{SteEssDel03}, where spectral partitioning has been
applied to \sam, or they can emerge from the underlying 3D structure of the
problem, as described in \cite{NiSteDel07}. The computational gain of such
methods is obtained by limiting the combinatorial explosion of the algorithm
complexity as the number of images and points increases.

The second strategy is to select a subset of the input images and points
that subsumes the entire solution. Hierarchical sub-sampling was pioneered by
\cite{FitZis98}, using a balanced tree of trifocal tensors over a video
sequence. The approach was subsequently refined by \cite{Nis00}, adding
heuristics for redundant frames suppression and tensor triplet selection. In
\cite{ShuKeZha99} the sequence is divided into segments, which are resolved
locally. They are subsequently merged hierarchically, eventually using a
representative subset of the segment frames. A similar approach is followed in
\cite{GibEA02}, focusing on obtaining a well behaved segment subdivision and on
the robustness of the following merging step. The advantage of these methods
over their sequential counterparts lies in the fact that they improve error
distribution on the entire dataset and bridge over degenerate
configurations. In any case, they work for video sequences, so they cannot be applied
to unordered, sparse images. The approach of \cite{SnaSeiSze08} works with
sparse datasets and is based on selecting a subset of images whose model
provably approximates the one obtained using the entire set. This considerably
lowers the computational requirements by controllably removing redundancy from
the dataset. Even in this case, however, the images selected are processed
incrementally. Moreover, this method does not avoid computing the epipolar
geometry between all pairs of images.

 Within the solutions aimed at reducing the impact of the bundle
  adjustment phase, hierarchical approaches include
  \cite{SchZis02,Ni2012,GheFarFus10} and this paper. The first can be
considered as the first paper where the idea has been set forth: a spanning tree
is built to establish in which order the images must be processed. After that,
however, the images are processed in a standard incremental way. The approach
described in \cite{Ni2012} is based on recursive partitioning of the problem
into fully-constrained sub-problems, exploiting the bipartite structure of the
visibility graph. The partitioning operates on the problem variables, whereas
our approach works on the input images.

Orthogonally to the aforementioned approaches, a solution to the the
computational complexity of \sam is to throw additional computational power at
the problem \cite{AgaEA09}. Within such a framework, the former algorithmic
challenges are substituted by load balancing and subdivision of tasks. Such a
direction of research strongly suggest that the current monolithic pipelines
should be modified to accommodate ways to parallelize and optimally split the
workflow of \sam tasks.  In \cite{Frahm2010} image selection (via clustering) is
combined with a highly parallel implementation that exploits graphic processors
and multi-core architectures.

The impact of the bundle adjustment phase can also be  reduced by adopting a
different paradigm in which \emph{first} the motion is recovered and \emph{then}
the structure is computed. All these methods start from the relative orientation
of a subset of camera pairs (or triplets), computed from point correspondences,
then solve for the absolute orientation of all the cameras (\emph{globally}),
reconstruct 3D points by intersection, and finally run a single bundle
adjustment to refine the reconstruction. Camera internal parameters are
required.

The method described in \cite{Arie12} solves a homogeneous linear system based
on a novel decomposition of the essential matrix that involves the absolute
parameters only. In \cite{Snavely14} nonlinear optimization is performed to
recover camera translations given a network of both noisy relative translation
directions and 3D point observations. This step is preceded by outlier removal
among relative translations by solving simpler low-dimensional subproblems.  The
authors of \cite{disco11} propose a discrete Markov random field formulation in
combination with Levenberg-Marquardt minimization. This technique requires
additional information as input, such as geotag locations and vanishing points.
Other approaches (e.g. \cite{Kahl08, MartinecP07, SinhaSS10}) compute
translations together with the structure, involving a significant number of
unknowns. The method presented in \cite{Brand04} proposes a fast spectral
solution by casting translation recovery in a graph embedding problem.  Govindu
in \cite{Govindu01} derives a homogeneous linear system of equations in which
the unknown epipolar scaling factors are eliminated by using cross products, and
this solution is refined through iterative reweighted least squares.  The
authors of \cite{Jiang13} propose a linear algorithm based on an approximate
geometric error in camera triplets. Moulon et al. \cite{Moulon13} extract
accurate relative translations by using an a-contrario trifocal tensor
estimation method, and then recover simultaneously camera positions and scaling
factors by using an $\ell_{\infty}$-norm approach. Similarly to \cite{Jiang13},
this method requires a graph covered by contiguous camera triplets.  The authors
of \cite{Basri14} propose a two-stage method in which relative translation
directions are extracted from point correspondences by using robust subspace
optimization, and then absolute translations are recovered through semidefinite
programming.

Another relevant issue in \sam is the level of generality, i.e., the number of
assumption that are made concerning the input images, or, equivalently the
amount of extra information that is required in addition to pixel
values. Existing pipelines either assume known internal parameters
\cite{BroLow05,IrsZacBis07,MartinecP07,EnqvistKO11,OlssonE11a,Arie12,Moulon13},
or constant internal parameters \cite{VerGoo06}, or rely on EXIF data plus
external information (camera CCD dimensions)
\cite{KamEA06,SnaSeiSze06}. Methods working in large scale environments usually
rely on a lot of additional information, such as camera calibration and GPS/INS
navigation systems \cite{CorEA08,PolEA08} or geotags \cite{disco11}.

\subsection{Autocalibration: related work.}
Autocalibration (a.k.a.~self-calibration) has generated a lot of theoretical
interest since its introduction in the seminal paper by Maybank and Faugeras
\cite{MayFau92}.
The attention created by the problem however is inherently practical, since it
eliminates the need for off-line calibration and enables the use of content
acquired in an uncontrolled setting.
Modern computer vision has partly sidestepped the issue by using ancillary
information, such as {EXIF} tags embedded in some image formats. Unfortunately
it is not always guaranteed that such data will be present or consistent with
its medium, and do not eliminate the need for reliable autocalibration
procedures.

A great deal of published methods rely on equations involving the dual image of
the absolute quadric (DIAQ), introduced by Triggs in \cite{Tri97}. Earlier
approaches for variable focal lengths were based on linear, weighted systems
\cite{PolKocGoo98,PolVerVan02}, solved directly or iteratively \cite{HeyCip01}.
Their reliability has been improved by more recent algorithms, such as
\cite{Cha07}, solving super-linear systems while directly forcing the positive
definiteness of the DIAQ.
Such enhancements were necessary because of the structural non-linearity of the
task: for this reason the problem has also been approached using branch and
bound schemes, based either on the Kruppa equations \cite{FusBenFarBus04}, dual
linear autocalibration \cite{Bar07} or the modulus constraint \cite{Cha07b}.

The algorithm described in \cite{HarHayAgaRei99} shares, with the branch and
bound approaches, the guarantee of convergence; the non-linear part,
corresponding to the localization of the plane at infinity, is solved
exhaustively after having used the cheiral inequalities to compute explicit
bounds on its location.

\subsection{Overview}

\begin{figure*}[htbp]
\centering \includegraphics[width=1.0\linewidth]{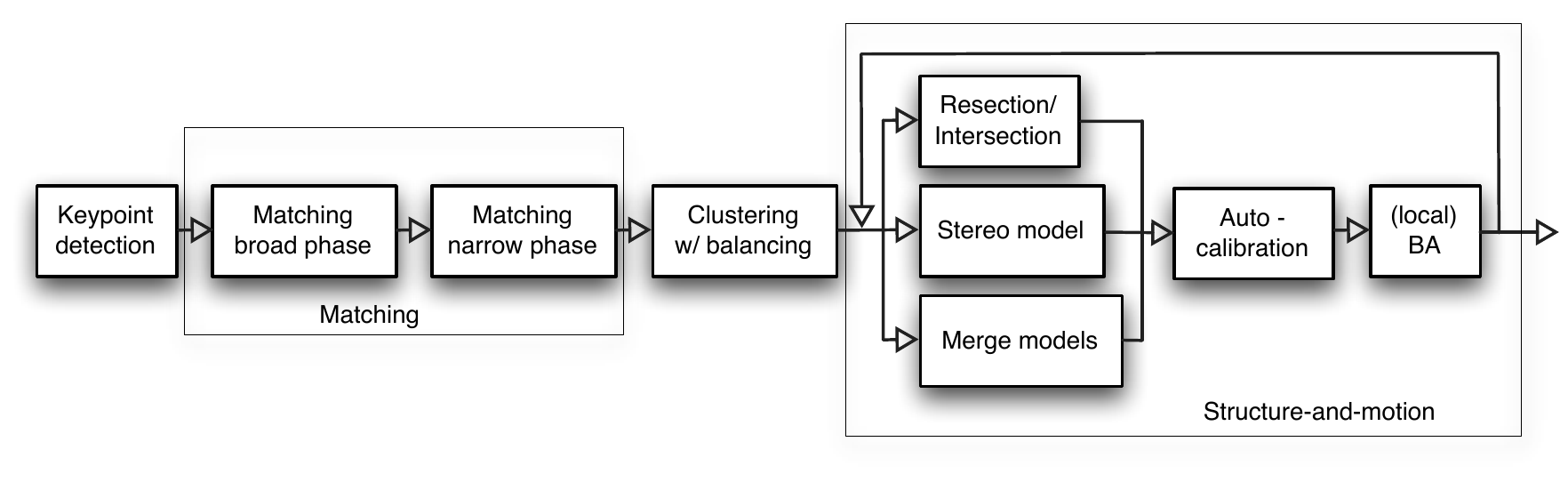}
\caption{Simplified overview of the pipeline. The cycle inside the \sam module
  corresponds to dendrogram traversal. Autocalibration can be switched
  on/off depending on circumstances. Bundle adjustment (BA) can be global or
  local and include, or not, the internal parameters.
\label{fig:schema}}
\end{figure*}

This paper describes a hierarchical and parallelizable scheme for \sam;
please refer to Fig.~\ref{fig:schema} for a graphical overview. The front end of
the pipeline is keypoint extraction and matching (Sec.~\ref{sec:match}), where
the latter is subdivided into two stages: the first (``broad phase'') is devoted
to discovering the tentative topology of the epipolar graph, while the second
(``narrow phase'') performs the fine matching and computes the epipolar geometry.

Images are then organized into a dendrogram by clustering them according to
their overlap (Sec.~\ref{sec:dendro}).  A new 
clustering strategy, derived from
the simple linkage, is introduced (Sec.~\ref{sec:balance}) that makes the
dendrogram more balanced, thereby approaching the best-case complexity of the
method.

The \sam computation proceeds hierarchically along this tree, from the leaves to
the root (Sec.~\ref{sec:hierarchical}).  Images are stored in the leaves,
whereas partial models correspond to internal nodes.  According to the type of
node, three operations are possible: stereo modeling (image-image),
resection-intersection (image-model) or merging (model-model). Bundle adjustment
is run at every node, possibly in its ``local'' form (Sec.~\ref{sec:localba}).

We demonstrate that this paradigm has several advantages over the
sequential one, both in terms of computational performance (which improves by
one order of magnitude on average) and overall error containment.

Autocalibration (Sec.~\ref{sec:urec}) is performed on partial models during the
dendrogram traversal.  First, the location of the plane at infinity is derived
given two perspective projection matrices and a guess on their internal
parameters, and subsequently this procedure is used to iterate through the space
of internal parameters looking for the best collineation that makes the
remaining cameras Euclidean.
This approach has several advantages: it is fast, easy to implement and reliable,
since a reasonable solution can always be found in non-degenerate
configurations, even in extreme cases such as when autocalibrating just two
cameras.

Being conscious that ``the devil is in the detail'', Section \ref{sec:devil}
reports implementation details and heuristics for setting the parameters of the
pipeline.

The experimental results reported in Sec.~\ref{sec:exp} are exhaustive and
analyze the output of the pipeline in terms of accuracy, convergence and
speed.

We report here on the  latest version of our pipeline, called
\sammy. Previous variants have been described in \cite{FarFusGhe09} and
\cite{GheFarFus10} respectively. The main improvements are in the matching phase
and in the autocalibration that now integrates the method described in
\cite{GheFus10}. The geometric stage has been carefully revised to make it more
robust, to the point where -- in some cases -- bundle adjustment is not needed
any more except at the end of the process.  Most efforts have been made in the
direction of a robust and automatic approach, avoiding unnecessary parameter
tuning and user intervention.

\section{Keypoint detection and  matching \label{sec:match}}

In this section we describe the stage of \sammy that is devoted to the automatic
extraction and matching of keypoints among all the $n$ available images.  Its
output is to be fed into the geometric stage, that will perform the actual \sam
recovery. Good and reliable matches are the key for any geometric computation.

Although most of the building blocks of this stage are fairly standard
techniques, we carefully assembled a procedure that is fully automatic, robust
(matches are pruned to discard as many outliers as possible) and computationally
efficient. The procedure for recovering  the epipolar graph is indeed  new.

\subsection{Keypoint detection.}

First of all, keypoints are extracted in all $n$ images.  
We implemented the keypoint detector proposed by \cite{Lin98}, where blobs with
associated scale levels are detected from scale-space extrema of the
scale-normalized Laplacian:
\begin{equation} 
\nabla^2_{norm} L(x, y, s) = 
s \, \nabla^2 \left(g(x, y; s) * f(x, y)\right) .
\end{equation} 
We used a 12-level scale-space and in each level the Laplacian is computed by
convolution (in CUDA) with a $3 \times 3$ kernel.

As for the descriptor, we implemented a 128-dimen\-sional radial descriptor
(similar to the log-polar grid of GLOH \cite{MS05}), based on the accumulated
response of steerable derivative filters. This combination of
detector/de\-scrip\-tor performs in a comparable way to SIFT and avoids patent
issues.

 Only a given number of keypoints with the strongest response
  overall are retained.  This number is a multiple of $n$, so as to fix the
  average quota of keypoints per image (details in Sec.~\ref{sec:devil}).

\subsection{Matching: broad phase.}\label{sec:broad}

As the images are unordered, the first objective is to recover the \emph{epipolar
  graph}, i.e., the graph that tells which images overlap (or can be matched)
with each other.

This must be done in a computationally efficient way, without trying to match
keypoints between every image pair. As a matter of fact, keypoint matching is
one of the most expensive stages, so one would like to reduce the number of images
to be matched from $O(n^2)$ to $O(n)$.

In this broad phase we consider only a small constant number of
  descriptors for each image.  In particular, we consider the keypoints with the
  highest scales, since their descriptors are more representative of the whole
  image content.

Then, each keypoint descriptor is matched to its approximate nearest neighbors
in feature space, using the ANN library \cite{MouAry07} (with $\epsilon = 0.5$).
A 2D histogram is then built that registers in each bin the number of matches
between the corresponding images.

Consider the complete weighted graph $G=(V,E)$ where $V$ are images and the
weighted adjacency matrix is the 2D histogram. This graph is -- in general --
dense, having $\abs{V} = O(n^2)$. The objective is to extract a subgraph $G'$
with a number of edges that is linear in $n$.

In the approach of \cite{Brolow03}, also followed in \cite{FarFusGhe09}, every
image is connected (in the epipolar graph) to the $m$ images that have the
greatest number of keypoint matches with it.  This creates a graph with $O(mn)$
edges, where the average degree is $O(m)$ (by the handshaking lemma).

When the number of images is large, however, it tends to create cliques of very
similar images with weak (or no) inter-clique connections. On the other hand,
one would like to get an epipolar graph that is strongly connected, to avoid
over-fragmentation in the subsequent clustering phase. This idea is captured by
the notion of \emph{$k$-edge-connectedness}: In graph theory, a graph is
$k$-edge-connected if it remains connected whenever fewer than $k$ edges are
removed. So, the graph produced by the original approach has a low $k$, while
one would like to have $k$ as high as possible (ideally $k=m$), with same edge
budget.

We devised a strategy that builds a subgraph $G'$ of $G$ which is ``almost''
$m$-edge-connected by construction.
\begin{enumerate}
\item Build the maximum spanning tree of $G$: the tree is composed of $n-1$
  edges;
\item remove them from $G$ and add them to $G'$; 
\item repeat  $m$ times. 
\end{enumerate}
In the hypothesis that $m$ spanning trees can be extracted from $G$, the
algorithm produces a subgraph $G'$ that is $m$-edge-connected (a simple proof of
this is given in \ref{sec:app2}).  Please note that by taking the \emph{maximum}
spanning tree we favor edges with high weight. So this strategy can be seen as a
compromise between picking pairs with the highest score in the histogram, as in
the original approach, and creating a strongly connected epipolar graph.

If the hypothesis about $G$ is not verified, a spanning forest will be obtained at
a certain iteration, and $G'$ will not be $m$-edge-connected. However, when $m
\ll \abs{E}$ one could expect that ``almost'' $m$ spanning trees can be
extracted from $G$ without disconnecting it.

\subsection{Matching: narrow phase.}

Keypoint matching follows a nearest neighbor approach \cite{Low04}, with
rejection of those keypoints for which the ratio of the nearest neighbor
distance to the second nearest neighbor distance is greater than a threshold
(see Sec.~\ref{sec:devil}). 
Matches that are not injective are discarded.

 In order to speed up the matching phase we employ a keypoint
 clustering technique similar to \cite{alhwarin2010vf}. Every
 keypoint is associated with a different cluster according to its dominant angle,
as recorded in the descriptor.
 Only keypoints belonging to the same cluster are matched
  together (in our implementation we used
  eight equidistant angular clusters): this breaks down the quadratic complexity of the matching phase at
  the expense of loosing some matches at the border of the clusters.

Homographies and fundamental matrices between pairs of matching images are then
computed using M-estimator SAmple Consensus (MSAC) \cite{TorZis00}, a variation
of RANSAC that gives outliers a fixed penalty but scores inliers on how well
they fit the data. This makes the output less sensitive to a higher inlier
threshold, thereby rendering less critical the choice of the threshold, at no
extra computational cost with respect to RANSAC. The random sampling is done
with a \emph{bucketing} technique \cite{Zha95}, which forces keypoints in the
sample to be spatially separated. This helps to reduce the number of iterations
and provide more stable estimates.  Since RANSAC and its variants (like MSAC)
have a low statistical efficiency, the model must finally be re-estimated on
a refined set of inliers\footnote{It is understood that when we refer to
  MSAC in the following, this procedure is always carried out.}.

Let $e_i$ be the residuals of \emph{all} the $N$ keypoints after MSAC, and let
$S^*$ be the sample that attained the best score; following \cite{Ste99}, a
robust estimator of the scale is:
\begin{equation}
\sigma^* = 1.4826 \left( 1 + \frac{5}{N-\abs{S^*}} \right) \sqrt{\med_{i\not \in
    S^* } e_i^2}.
\end{equation}
The resulting set of inliers are those points such that
\begin{equation} \abs{e_i}  < \theta \sigma^*, \end{equation}
where $\theta$ is a constant (we used 2.5).

The model parameters are re-estimated on this set of inliers via
least-squares minimization of the (first-order approximation of the) geometric
error \cite{LuoFau96,ChuPajStu05}.

The more likely model (homography or fundamental matrix) is selected according
to the Geometric Robust Information Criterion (GRIC) \cite{Tor97}:
\begin{eqnarray}
\mathrm{GRIC} & = & \sum \rho(e_i^2) + nd \log(r) + k \log (rn) \\ 
\nonumber \rho(x) & = & \min \left( x/{\sigma^2}, 2(r-d) \right)
\end{eqnarray}
where $\sigma$ is the standard deviation of the measurement error, $k$ is the number
of parameters of the model, $d$ is the dimension of the fitted manifold, and $r$ is
the dimension of the measurements. In our case, $k = 7, d = 3, r = 4$ for
fundamental matrices and $k = 8, d = 2, r = 4$ for homographies. The model with
the lower GRIC is the more likely.

\begin{figure*}[bhtp]
\centering \includegraphics[width=1.0\linewidth]{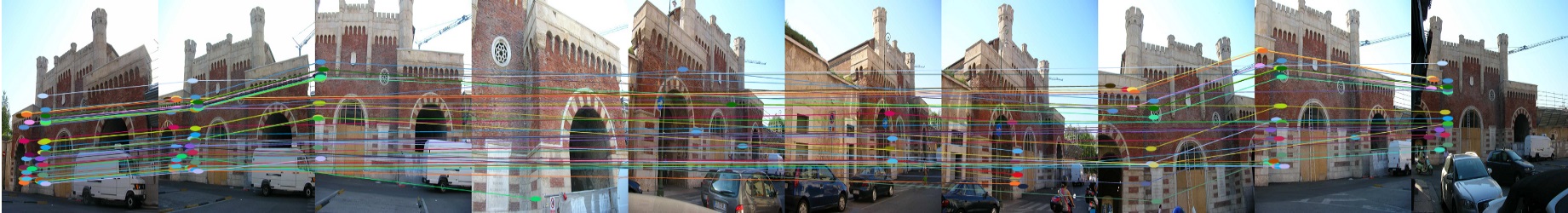}
\caption{Tracks over a 12-images set. For the sake of readability only a sample
  of 50 tracks over 2646 have been plotted.
\label{fig:tracce}}
\end{figure*}

 In the end, if the number of remaining matches $n_i$ between two
  images is less than 20\% of the total number of matches before MSAC, then they
  are discarded.  The rationale is that if an excessive fraction of oultliers
  have been detected, the original matches are altogether unreliable. A similar
  formula is derived in \cite{Brolow03} on the basis of a probabilistic model. As a
  safeguard, a threshold on the minimum number of matches is also set (details
  in Sec.~\ref{sec:devil}).

After that, keypoint matching in multiple images are connected into
\emph{tracks} (see Figure \ref{fig:tracce}): consider the undirected graph $G =
(V,E)$ where $V$ are the keypoints and $E$ represents matches; a track is a
connected component of $G$.  Vertices are labeled with the image the keypoints
belong to: an inconsistency arises when in a track a label occurs more than once.
Inconsistent tracks and those shorter than three frames are discarded
\footnote{There is nothing in principle that prevents the pipeline from
  working also with tracks of length two. The choice of cutting these tracks is a
  heuristics aimed at removing little reliable correspondences.}.  A track
represents the projection of a single 3D point imaged to multiple exposures;
such a 3D point is called a \emph{tie-point}.

\section{Clustering images \label{sec:dendro}}

The images are organized into a tree with agglomerative clustering, using a
measure of overlap as the distance. The \sam computation then follows this tree
from the leaves to the root. As a result, the problem is broken into smaller
instances, which are then separately solved and combined.

Algorithms for image  clustering have been proposed in the literature in the
context of \sam \cite{SchZis02}, panoramas \cite{Brolow03}, image mining
\cite{QuaLeiGoo08} and scene summarization \cite{SimSnaSei07}. The distance
being used and the clustering algorithm are application-specific.

In this paper we deploy an image affinity measure that befits the \sam task. It
is computed by taking into account the number of tie-points visible in both
images and how well their projections (the keypoints) spread over the images. In
formulae, let $S_i$ and $S_j$ be the set of  visible tie-points in
image $I_i$ and $I_j$ respectively:
\begin{equation}
  a_{i,j} = \frac{1}{2}\frac{\abs{S_i \cap S_j}}{\abs{S_i \cup S_j}} + 
\frac{1}{2} \frac{CH(S_i)+CH(S_j)}{A_i+A_j} 
\label{eq:affinity} 
\end{equation}
where $CH(\cdot)$ is the area of the convex hull of a set of image points and $A_i$
($A_j$) is the total area of image $I_i$ ($I_j$).  The first term is an affinity
index between sets, also known as the Jaccard index. The distance is $(1 -
a_{i,j})$, as $a_{i,j}$ ranges in $[0,1]$.

The general agglomerative clustering algorithm proceeds in a bottom-up manner:
starting from all singletons, each sweep of the algorithm merges the two
clusters with the smallest distance between them. 
The way the distance between clusters is computed produces different flavors of
the algorithm, namely the simple linkage, complete linkage and average linkage
\cite{DudHar73}. We selected the \emph{simple linkage} rule: The distance
between two clusters is determined by the distance of the two closest objects
(nearest neighbors) in the different clusters.  

Simple linkage clustering is appropriate to our case because: i) the clustering
problem \emph{per se} is fairly easy, ii) nearest neighbors information is
readily available with ANN and iii) it produces ``elongated'' or ``stringy''
clusters which fits very well with the typical spatial arrangement of images
sweeping a certain area or building.

\section{Hierarchical \sam \label{sec:hierarchical}}

Before describing our hierarchical approach, let us set the notation and review
the geometry tools that are needed. A \emph{model} is a set of cameras and 3D
points expressed in a local reference frame (\emph{stereo-model} with two
cameras).  The procedure of computing 3D point coordinates from corresponding
points in multiple images is called \emph{intersection} 
  (a.k.a.~triangulation). Recovering the camera matrix (fully or limited to the
external parameters) from known 3D-2D correspondences is called
\emph{resection}. The task of retrieving the relative position and attitude of
two cameras from corresponding points in the two images is called \emph{relative
  orientation}. The task of computing the rigid (or similarity) transform that
brings two models that share some tie-points into a common reference frame is
called \emph{absolute orientation}.

Let us assume {\it pro tempore} that the internal parameters are known; this
constraint is removed in Sec.~\ref{sec:urec}.

\bigskip

\begin{figure}[tbhp]
\centering \includegraphics[width=1.0\linewidth]{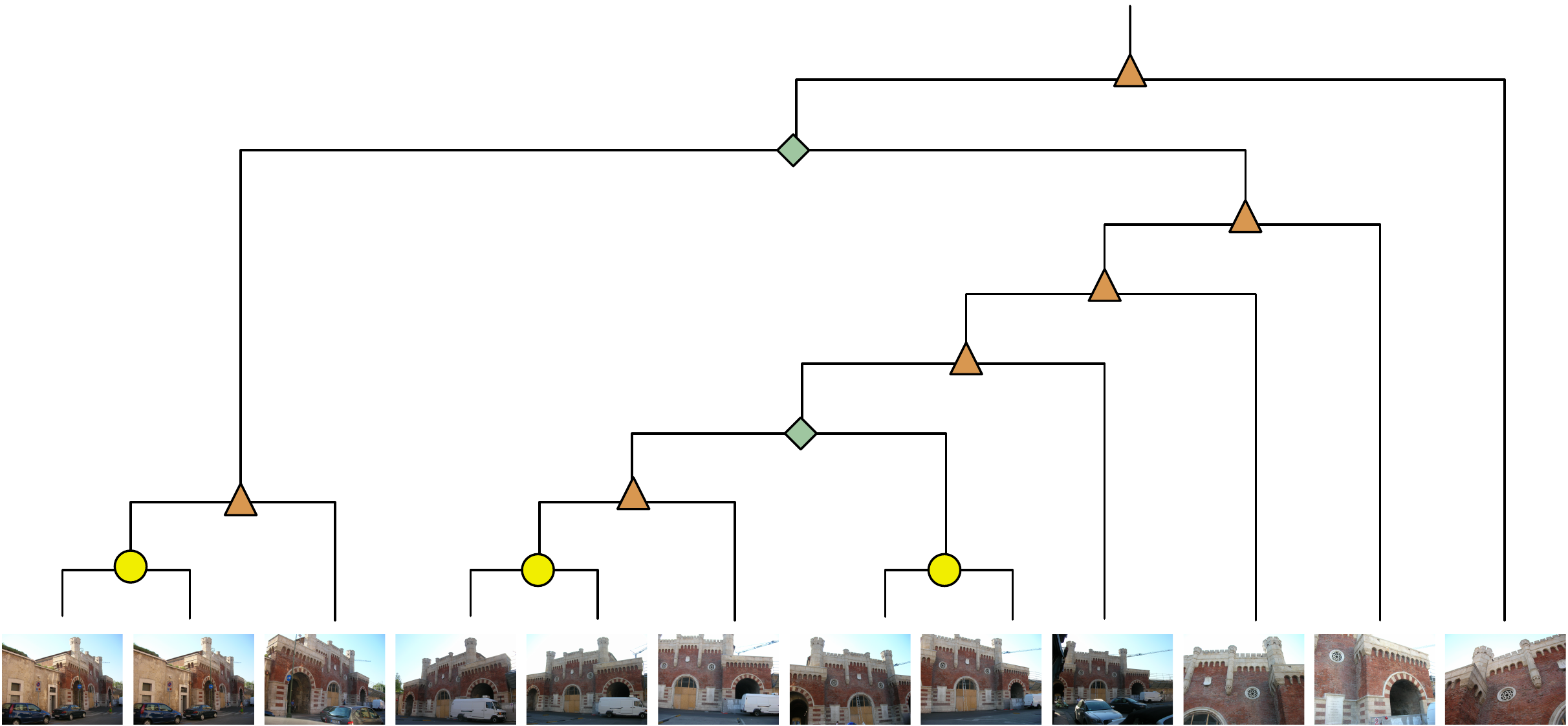}
\caption[foo]{An example of a  dendrogram for a 12 image set. The circle
  \includegraphics[height=1.5ex]{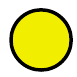} corresponds to the creation
  of a stereo-model, the triangle
  \includegraphics[height=1.5ex]{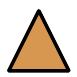} corresponds to a
  resection-intersection, the diamond
  \includegraphics[height=1.5ex]{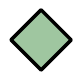} corresponds to a fusion of
  two partial independent models.}
\label{fig:2.1}
\end{figure}

Images are grouped together by agglomerative clustering, which produces a
hierarchical, binary cluster tree, called a \emph{dendrogram}.
Every node in the tree represents a partial, independent model. From the
processing point of view, at every node in the dendrogram an action is taken
that augments the model, as shown in Figure \ref{fig:2.1}.

Three operations are possible: When two images are merged a stereo-model is
built (relative orientation + intersection). When an image is added to a cluster
a resection-intersection step is taken (as in the standard sequential
pipeline). When two non-trivial clusters are merged, the respective models must
be conflated by solving an absolute orientation problem (followed by
intersection). Each of these steps is detailed in the following.

While it is useful to conceptually separate the clustering from the modeling,
 the two phases actually occur simultaneously: during the simple linkage
iteration, every time a merge is attempted the corresponding modeling action is
taken. If it fails, the merge is discarded and the next possible merge is
considered.

\subsection{Stereo-modeling.}

The parameters of the relative orientation of two given cameras are obtained by
factorization of the essential matrix \cite{Har92}. This is 
  equivalent to knowing the external parameters of the two cameras in a local
reference frame, and since the internal parameters are already known, the two
camera matrices are readily set up.  Then tie-points are obtained by
\emph{intersection} (see Sec.~\ref{sec:inter} ahead) from the tracks involving
the two images, and the model is refined with bundle adjustment \cite{bundle}.

\bigskip

It is worth noting that in order for this stereo-modeling to be successful the
two images must satisfy two conflicting requirements: to have both a large
number of tie-points in common and a baseline sufficiently large so as to allow
a well-conditioned solution.  The first requirement is implemented by the
affinity defined in \eqref{eq:affinity}, but the second is not considered; as a
result, the pairing determined by image clustering is not always the best choice
as far as the relative orientation problem is concerned. Since in our pipeline
the clustering and the \sam processing occurs simultaneously, these pairs will
be discarded by simple sanity-checks before and after attempting to perform the
stereo-modeling. The a-priori check requires that the relationship between the
two images is described by a fundamental matrix (instead of a homography),
according to the GRIC score. The a-posteriori check considers the residual error
and the cheirality check of the points before and after the bundle adjustment.

\subsection{Intersection.} \label{sec:inter}

Intersection (a.k.a.~triangulation) is performed by the iterated linear LS method
\cite{HarStu97}.  Points are pruned by analyzing the condition number of the
linear system and the reprojection error.  The first test discards
ill-conditioned intersections, using a threshold on the condition number of the
linear system ($10^4$, in our experiments). The second test applies the
so-called X84 rule\footnote{This rule is consistently used in the following
  stages to set data-dependent thresholds whenever required.}
\cite{HamRouRonSta86}, that establishes that, if $e_i$ are the residuals, the
inliers are those points such that
\begin{equation} | e_i - \med_j e_j| < 5.2 \med_i| e_i - \med_j e_j| .
 \end{equation}

A safeguard threshold on the reprojection error is also set (details in
Sec.~\ref{sec:devil}).

In general, the intersection module obeys the following strategy. As soon as one
track reaches length two in a given model (i.e. at least two images of the track
belongs to the model), the coordinates of the corresponding tie-point are
computed by intersection. If the operation fails (because of one of the sanity
checks described above) the 3D point is provisionally discarded but the track is
kept.  An attempt to compute the tie-point coordinates is undertaken every time
the length of the track increases within the model.

\subsection{Resection}
The tie-points belonging to the model that are also visible in the image to be
added provides a set of 3D-2D correspondences, that are exploited to glue the
image to the partial model. This is done by resection, where only the external
parameters of the camera are to be computed (a.k.a.~external orientation
problem). We used the PPnP algorithm described in \cite{GarCroFus12} inside
MSAC, followed by non-linear minimization of the reprojection error at the end.

After resection, which adds one image to the model, tie-points are updated by
intersection, and bundle adjustment is run on the resulting model.

\subsection{Merging two models.}
When two partial independent (i.e., with different reference systems) models are
are to be conflated into one, the first step is to register one onto the other
with a similarity transformation. The common tie-points are used to solve an
absolute orientation (with scale) problem with MSAC.

Given the scale ambiguity, the inlier threshold for MSAC is hard to set. In
\cite{RaguramF11} a complex technique for the automatic estimation of the inlier
threshold in 3D is proposed. We take a simpler but effective approach: instead
of considering the length of the 3D segments that connect corresponding points
as the residuals, we look at the average length of their 2D projections in the
images; in this way a meaningful inlier threshold in pixels can be set easily.
The final transformation, computed with the Orthogonal Procrustean (OP) method
\cite{Kan93,ShoCar70}, minimizes the proper geometric residual, i.e. the sum of
squared distances of 3D points.

Once the models are registered, tie-points are updated by intersection, and the
new model is refined with bundle adjustment.

\bigskip

This hierarchical algorithm can be summarized as follows:
\begin{enumerate}
\item Solve many independent relative orientation problems at the leaves of the
  tree, producing many independent stereo-models.

\item Traverse the tree; in each node one of these operations takes place:
\begin{enumerate}

\item Update one model by adding one image with resection followed by
  intersection;

\item Merge two independent models with absolute orientation.

\end{enumerate}

\end{enumerate}

Steps 1.~and 2.(a) are the resection-intersection steps of classical sequential
pipelines, as Bundler.  Step 2.(b) summons up the photogrammetric Independent
Models Block Adjustment (IMBA) \cite{Kra97}, where for each pair of overlapping
photographs a stereo-model is built and then all these independent models are
simultaneously transformed into a common reference frame with absolute
orientation.

If the tree reduces to a chain, the algorithm is the sequential one, whereas if
the tree is perfectly balanced, only steps 2.(b) are taken, and the resulting
procedure resembles the IMBA, besides the fact that our models are disjoint and
that models are recursively merged in pairs.

Compared to the standard sequential approach, this framework has a lower
computational complexity, is independent of the initial pair of images, and
copes better with drift problems, typical of sequential schemes.

\subsection{Complexity analysis.}
The hierarchical approach that has been outlined above allows us to decrease the
computational complexity with respect to the sequential \sam pipeline.  Indeed,
if the number of images is $n$ and every image adds a constant number of tie-points
$\ell$ to the model, the computational complexity\footnote{We are
  considering here only the cost of bundle adjustment, which clearly dominates
  the other operations.} in time of sequential \sam is $O(n^5)$, whereas the
complexity of \sammy (in the best case) is $O(n^4)$.

The cost of bundle adjustment with $m$ tie-points and $n$ images is $O(mn (m+2n)^2)$
\cite{ShuKeZha99}, hence it is $O(n^4)$ if $m = \ell n$.

In sequential \sam, adding image $i$ requires a constant number of bundle
adjustments (typically one or two) with $i$ images, hence the complexity is
\begin{equation}
\sum_{i=1}^n O(i^4) = O(n^5).
\end{equation} 
In the case of the hierarchical approach, consider a node of the dendrogram
where two models are merged into a model with $n$ images.  The cost $T(n)$ of
adjusting that model is given by $O(n^4)$ plus the cost of doing the same onto
the left and right subtrees. In the hypothesis that the dendrogram is well
balanced, i.e., the two models have the same number of images, this cost is given by $2
T(n/2) $. Hence the asymptotic time complexity $T$ in the best case is given by
the solution of the following recurrence:
\begin{equation}
T(n) = 2 T(n/2) + O(n^4)
\end{equation} that is $T(n) = O(n^4)$ by the third branch of the Master's theorem
\cite{CorLeiRiv01}.

The worst case is when a single model is built up by adding one image at a time.
In this case, which corresponds to the sequential case, the dendrogram is
extremely unbalanced and the complexity drops to $O(n^5)$.

\section{Dendrogram balancing \label{sec:balance}}

As demonstrated in precedence, the hierarchical framework can provide a provable
computational gain, provided that the resulting tree is well-balanced.  The
worst case complexity, corresponding to a sequence of single image additions, is
no better than the standard sequential approach. It is therefore crucial to
ensure a good balance during the clustering phase. Our solution is to employ a
novel clustering procedure, which promotes the creation of better balanced
dendrograms.

The image clustering procedure proposed in the previous section allows us to
organize the available images into a hierarchical cluster structure (a tree)
that will guide the \sam process. This approach decreases the computational
complexity with respect to sequential \sam pipelines, from $O(n^5)$ to $O(n^4)$
in the best case, i.e.  when the tree is well balanced ($n$ is the number of
images). If the tree is unbalanced this computational gains vanishes. It is
therefore crucial to enforce the balancing of the tree.

 The preceding solution, which used the simple rule, specified that the distance
 between two clusters is to be determined by the distance of the two closest
 objects (nearest neighbors) in the different clusters.  In order to produce
 better balanced trees, we modified the agglomerative clustering strategy as
 follows: starting from all singletons, each sweep of the algorithm merges the
 pair with the smallest cardinality among the $\ell$ closest pair of clusters.  The
 distance is computed according to the simple linkage rule. The cardinality of a
 pair is the sum of the cardinality of the two clusters.  In this way we are
 softening the “closest first” agglomerative criterion by introducing a
 competing “smallest first” principle that tends to produce better balanced
 dendrograms. 

The amount of balancing is regulated by the parameter $\ell$: when $\ell = 1$
this is the standard agglomerative clustering with no balancing; when $\ell \ge
n/2$ ($n$ is the number of images) a perfect balanced tree is obtained, but the
clustering is poor, since distance is largely disregarded.

\begin{figure*}[t]
 \centering 
\includegraphics[width=0.495\linewidth]{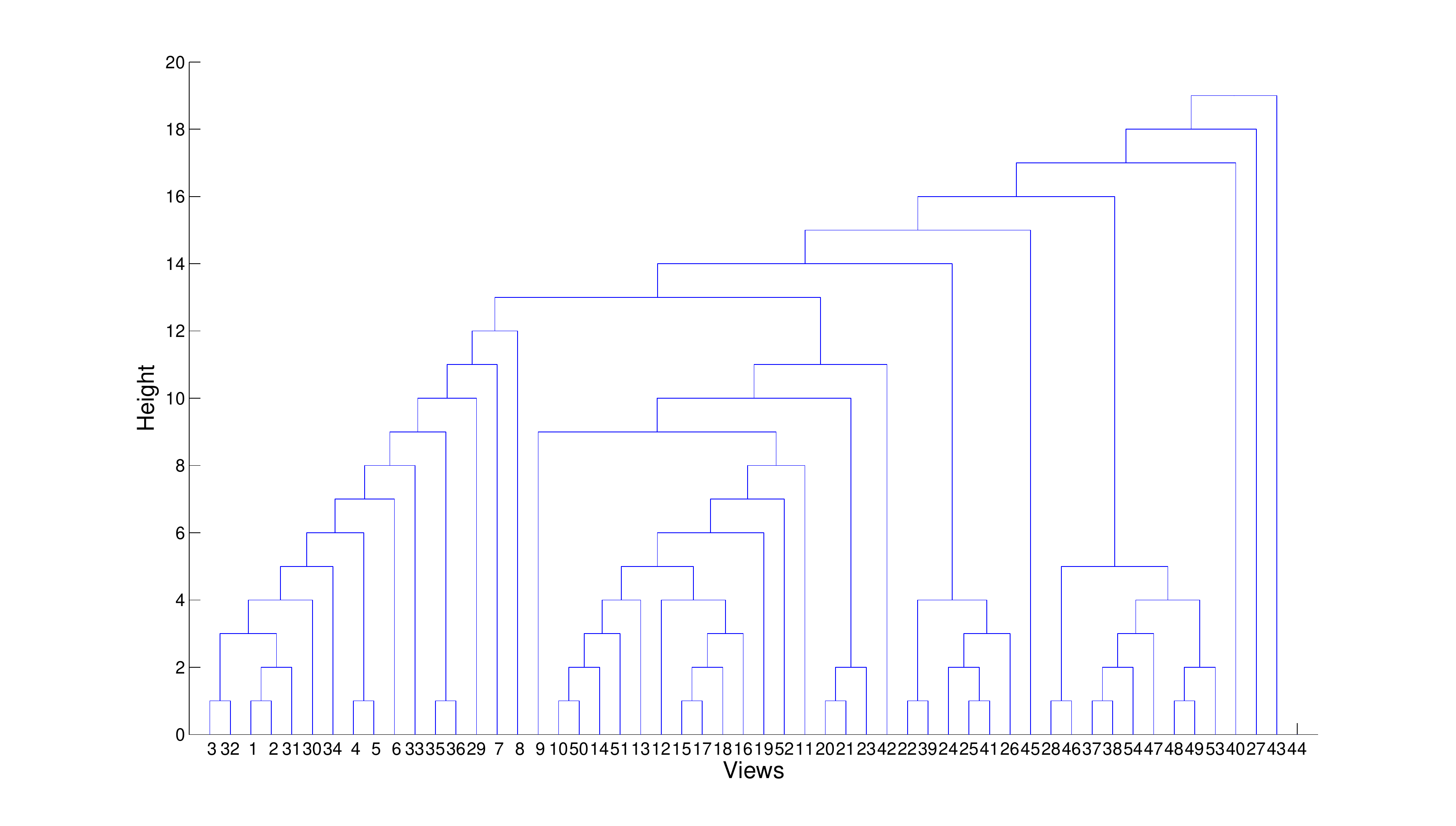}
\includegraphics[width=0.495\linewidth]{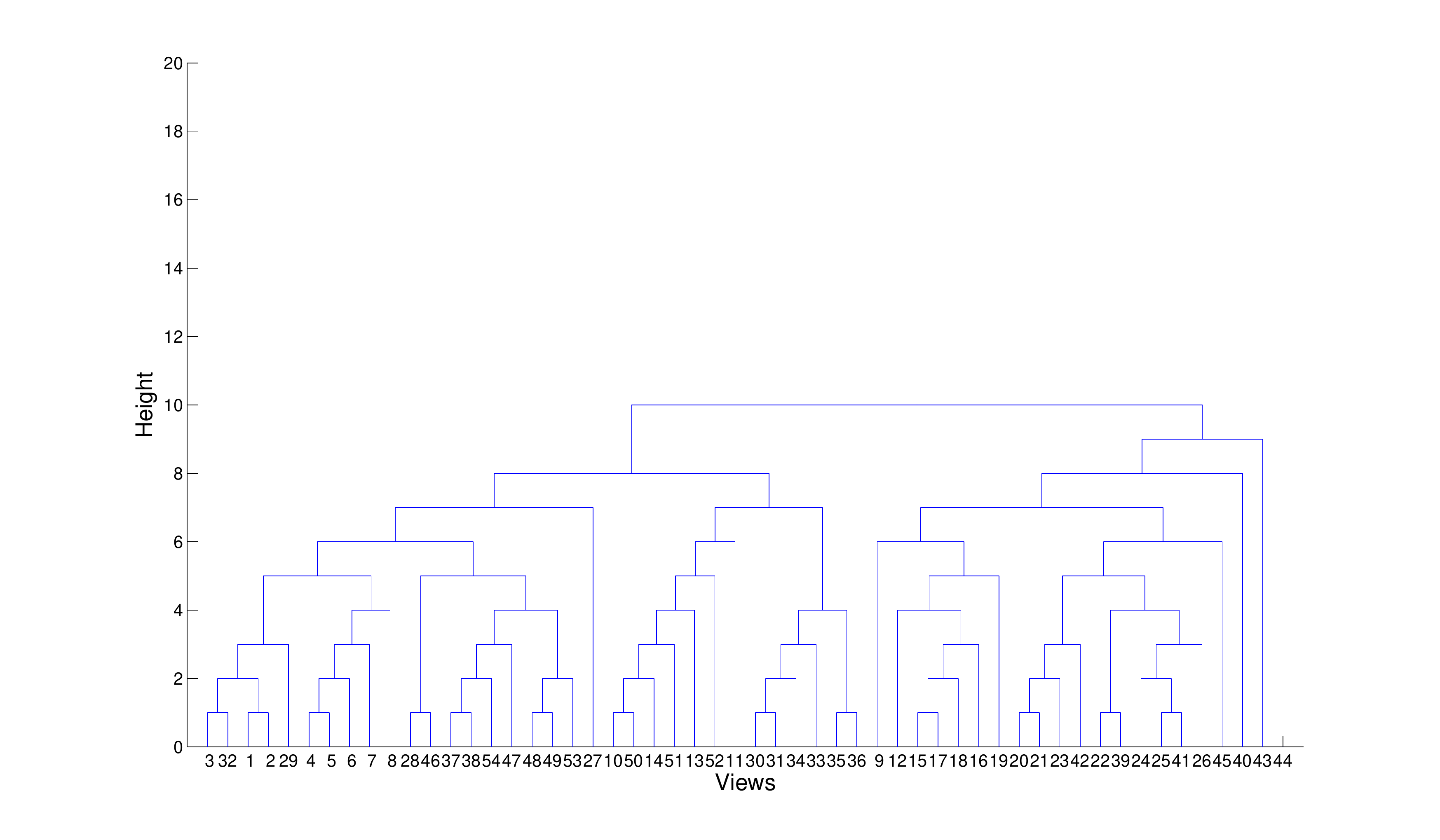}
 \caption{An example of the dendrogram produced by simple linkage (left) and the
   balanced rule on a 52-images set. An example of the dendrogram produced by
   \cite{FarFusGhe09} (left) and the more balanced dendrogram produced by our
   technique (right) on a 52-images set, with $\ell = 5$.
 \label{fig:2.9}}
 \end{figure*}

Figure \ref{fig:2.9} shows an example of balancing achieved by our
technique. The height of the tree is reduced from 14 to 9 and more initial pairs
are present in the dendrogram on the right.

\section{Local bundle adjustment}\label{sec:localba}

In the pursuit of further complexity reduction, we adopted a strategy that
consists in reducing the number of images to be used in the bundle adjustment in
place of the whole model. This strategy is an instance of local bundle
adjustment \cite{ZhaSha03,MouEA06}, which is often used for video sequences,
where the active images are the most recent ones.  Let us concentrate on the
model merging step, as the resection is a special case of the latter. Consider
two models $A$ and $B$, where $A$ has fewer images than $B$. We always transform
the smallest onto the largest (if one is projective it is always the smallest).

The bundle adjustment involves all the images of $A$ and the subset of images of
$B$ that share some tracks with $A$ (tie-points that are visible in images in
both models). Let us call this subset $B'$. All the tie-points linking $B'$ and
$A$ are considered in the bundle adjustment.  Images in $B \setminus B'$ are not
moved by bundle adjustment but their tie-points are still considered in the
minimization in order to anchor $B'$ through their common tie-points. The
tie-points linking only cameras in $B \setminus B'$ are not considered.  This
strategy is sub-optimal because in a proper bundle adjustment all the images in
$B$ should be involved, even those that do not share any tie-point with $A$.
However, a bundle adjustment with all the images and all the
tie-points can be run  at the end to obtain the optimal solution.

\section{Uncalibrated hierarchical structure-and-motion \label{sec:urec}}

A model that differs from the true one by a projectivity is called
\emph{projective}. A model that differs from the true one by a similarity is
called Euclidean. The latter can be achieved when calibration parameters are
known, the former  can be obtained if images are uncalibrated.

In this section we relax the hypothesis that images are calibrated and
integrate the autocalibration algorithm in our pipeline, so that the resulting
model is still Euclidean.

The main difference from the procedure described in Sec.~\ref{sec:hierarchical}
is that now leaf nodes do not have proper calibration right from the start of
the \sam process.  The models is projective at the beginning, and as soon as one
reaches a sufficient number of images, the Euclidean upgrade procedure
(described in Section \ref{sec:autocal}) is triggered. Moreover, each step of
hierarchical \sam must be modified to accommodate for projective models, as
described in Sections \ref{sec:2vu}, \ref{sec:res-int}, and \ref{sec:merge}.

\bigskip

\subsection{Autocalibration \label{sec:autocal}}

Autocalibration starts from a projective model and seeks the
col\-li\-ne\-ation of space $H$ so as to transforms the model into a Euclidean
one.

Without loss of generality, the first camera of the Euclidean model can be
assumed to be $ P_1^{\scriptscriptstyle{E}} = \left[ K_1 \ \vert \ \mathbf{0}
  \right]$, so that the Euclidean upgrade $H$ has the following structure, since
$P_1^{\scriptscriptstyle{E}} = P_1 H$:
\begin{equation}
H = 
\left[
\begin{array}{cc}
    K_{1} & \mathbf{0} \\
    \mathbf{r}^{\top} & \lambda
\end{array}
\right]
\label{eq:Heuc}
\end{equation}
where $K_{1}$ is the calibration matrix of the first camera, $\mathbf{r}$ is  a
vector which determines the location of the plane at infinity and $\lambda$ is a
scale factor.

Our autocalibration technique is based on two
stages:
\begin{enumerate}
\item Given a guess on the internal parameters of two cameras compute a
  consistent upgrading collineation. This yields an estimate of all cameras but
  the first.
\item Score the internal parameters of these $n-1$ cameras based on the
  likelihood of skew, aspect ratio and principal point.
\end{enumerate}
The space of the internal parameters of the two cameras is enumerated and the
best solution is refined via non-linear least squares.

This approach has been introduced in \cite{GheFus10}, where it is compared with
several other algorithms  obtaining favorable results.

\subsubsection{Estimation of the plane at infinity.}
\label{sec:autocal1}

This section  describes a closed-form solution for the plane at infinity
(i.e., the vector $\mathbf{r}$) given two perspective projection matrices and
their internal parameters.

While the first camera is $P_1 = \left[ I \ \vert \ \mathbf{0} \right]$, the
second projective camera can be written as $ P_2 = \left[ A_2 \ \vert
  \ \mathbf{e}_2 \right]$, and its Euclidean upgrade is:
\begin{align}
P_2^{\scriptscriptstyle{E}} & =  K_2  \left[ R_2  \vert \mathbf{t}_2
  \right] \simeq P_2  H = \left[ A_2  K_1 + \mathbf{e}_2  \mathbf{r}^{\top}
  \vert  \lambda \mathbf{e}_2 \right] .
\end{align}
The rotation $R_2$ can therefore be equated to:
\begin{equation}
    R_2 \simeq K_2^{\scriptscriptstyle{-1}} \!\! \left( A_2 K_1 + \mathbf{e}_2
    \mathbf{r}^{\top} \right) = K_2^{\scriptscriptstyle{-1}} A_2 K_1 +
     K_2^{\scriptscriptstyle{-1}}  \mathbf{e}_2 \mathbf{r}^{\top}
\label{eq:rotequ}
\end{equation}

Using the constraints on orthogonality between rows or columns of a rotation
matrix, one can solve for $\mathbf{r}$ finding the value that makes the right
hand side of \eqref{eq:rotequ} equal to a rotation, up to a scale.

The solution can be obtained in closed form by noting that there always exists a
rotation matrix $R^*$ such as: $ R^* \mathbf{t}_2 = \left[ \Arrowvert
  \mathbf{t}_2 \Arrowvert \ 0 \ 0 \right]^{\top},$ where $\mathbf{t}_2 =
K_2^{\scriptscriptstyle{-1}} \mathbf{e}_2 $. Left multiplying it to
\eqref{eq:rotequ} yields:
\begin{equation}
    R^* R_2 \simeq  R^*  K_2^{\scriptscriptstyle{-1}} A_2  K_1
    + \left[ \Arrowvert \mathbf{t}_2 \Arrowvert \ 0 \ 0 \right]^{\top}
    \mathbf{r}^{\top}
\end{equation}

Calling $W = R^*  K_2^{\scriptscriptstyle{-1}} A_2  K_1$  and its rows $\mathbf{w}_i^{\top}$,
we arrive at the following:
\begin{equation}
	R^* R_2 = \left[
	\begin{array}{l}
		{\mathbf{w}_2}^{\top} + \Arrowvert \mathbf{t}_2 \Arrowvert \mathbf{r}^{\top} \\
		{\mathbf{w}_2}^{\top} \\
		{\mathbf{w}_3}^{\top} \\
	\end{array}
	\right] /  \Arrowvert \mathbf{w}_3 \Arrowvert
\label{eq:8}
\end{equation}
in which the last two rows are independent of the value of $\mathbf{r}$ and
the correct scale has been recovered normalizing  each side of the
equation to unit norm.

Since the rows of $ R^* R_2 $ are orthonormal, we can recover the first one
taking the cross product of the other two. Vector $\mathbf{r}$ is therefore
equal to:
\begin{equation}
    \mathbf{r} = \left( \mathbf{w}_2 \times \mathbf{w}_3  /  \Arrowvert
    \mathbf{w}_3 \Arrowvert - \mathbf{w}_1 \right)  /  \Arrowvert \mathbf{t}_2
    \Arrowvert
\label{eq:9}
\end{equation}
The upgrading collineation $H$ can be computed using \eqref{eq:Heuc}; the term
$\lambda$ can be arbitrarily chosen, as it will just influence the overall
scale.

When the calibration parameters are known only approximately, the right hand
side of \eqref{eq:8} is no more a rotation matrix.
However, \eqref{eq:9} will still yield the value of $\mathbf{r}$ that will
produce an \emph{approximate} Euclidean model.

\subsubsection{Estimation of the internal parameters.}

In the preceding section we showed how to compute the Euclidean upgrade $H$
given the calibration parameters of two cameras of the projective
model.

The autocalibration algorithm  loops through all possible
internal parameter matrices of two cameras $K_1$ and $K_2$, checking whether the
entire upgraded model has the desired properties in terms of $K_2
\ldots K_n$.  The process is well-defined, since the search space is naturally
bounded by the finiteness of the acquisition devices.

In order to sample the space of calibration parameters we can safely assume, as
customary, null skew and unit aspect ratio: this leaves the focal length and the
principal point location as free parameters. However, as expected, the value of
the plane at infinity is in general far more sensitive to errors in the
estimation of focal length values rather than the image center. Thus, we can
iterate just over focal lengths $f_1$ and $f_2$ assuming the principal point to
be centered on the image; the error introduced with this approximation is
normally well within the radius of convergence of the subsequent non-linear
optimization. The search space is therefore reduced to a bounded region of
$\mathbb{R}^2$.

To score each sample  $(f_1, f_2)$, we consider the aspect ratio, skew and
principal point location of the upgraded (i.e., transformed with $H$) camera
matrices and aggregate their respective value into a single cost function:
\begin{equation}
    \left\{ f_1,f_2 \right\} = \arg \min_{f_1,f_2} \sum^n_{\ell=2} \mathcal{C}^2(K_\ell)
\label{eq:sumCosts}
\end{equation}
where $K_\ell$ is the internal parameters matrix of the $\ell$-th camera after
the Euclidean upgrade determined by $(f_1, f_2)$, and ${C}(K)$ reflects the
degree to which $K$ meets a-priori expectations.  

Let us consider the \emph{viewport matrices} of the cameras, defined as:
\begin{equation}
    V = \frac{1}{2}
	\left[
	\begin{array}{ccc}
	    \sqrt{w^2 + h^2} & 0 & w \\
	    0 & \sqrt{w^2 + h^2} & h \\
		 0 & 0 & 2
	\end{array}
	\right] 
\label{eq:viewport}
\end{equation}
where $w$ and $h$ are respectively the width and height of each image.  Camera
matrices are normalized with $P \leftarrow V^{-1} P / \Arrowvert P_{3,1:3}
\Arrowvert $.  In this way, the principal point expected value is $(0,0)$ and
the focal range is $[1/3, 3]$. 
Therefore, the term of the cost function writes:
\begin{equation}
    \mathcal{C}(K) {=} \overbrace{w_{sk} | k_{1,2} |}^{\mathsf{skew}} {+}
    \overbrace{w_{ar} | k_{1,1} {-} k_{2,2} |}^{\mathsf{aspect\ ratio}} {+}
    \overbrace{w_{u_o} | k_{1,3} | {+} w_{v_o} |
      k_{2,3}|}^{\mathsf{principal\ point}}
\label{eq:costFunction}
\end{equation}
where $k_{i,j}$ denotes the entry $(i,j)$ of $K$ and $w$ are suitable weights,
computed as in \cite{PolVerVan02}. The first term takes into account the skew,
which is expected to be 0, the second one penalizes cameras with aspect ratio
different from 1 and the last two weigh down cameras where the principal point
is away from $(0,0)$.

Finally, the solution selected is refined by non-linear minimization of
Eq.~\eqref{eq:sumCosts}. Since it is usually very close to a minimum, just a few
iterations of a Levenberg-Marquardt solver are necessary for convergence.

\IncMargin{1em}
\begin{algorithm}
\DontPrintSemicolon
\SetCommentSty{textsf}
\BlankLine
\BlankLine
\SetKwInOut{Input}{input}
\SetKwInOut{Output}{output}
\Input{a set of PPMs $P$ and their viewports $V$}
\Output{their upgraded, Euclidean counterparts}
\BlankLine
\lForEach{$P$}
{
	$P \leftarrow V^{-1}   P   / \Arrowvert P_{3,1:3} \Arrowvert $
} \tcc*{normalization}
\BlankLine
\ForEach(\tcc*[f]{iterate over focal pairs}){$K_1,K_2$}
{
	\BlankLine
	\emph{compute $\Pi_\infty$} \;
	\emph{build $H$ from  \eqref{eq:Heuc}} \;
	\ForEach(\tcc*[f]{compute cost profiles}){$P$}
	{
		\BlankLine
		$P_{\scriptscriptstyle{E}} \leftarrow P   H$ \;
		$K \leftarrow$ \emph{internal of $P_{\scriptscriptstyle{E}}$} \;
		\emph{compute $\mathcal{C}(K)$ from \eqref{eq:costFunction}} \;
		\BlankLine
	}
	\BlankLine
}
\BlankLine
\emph{aggregate cost and select minimum} \;
\emph{refine non-linearly} \;
\BlankLine
\lForEach{$P$}
{
	$P \leftarrow V   P   H $
} \tcc*{de-normalization, upgrade}
\BlankLine
\caption{Autocalibration pseudo-code}
\label{alg:selfcal}
\end{algorithm}
\DecMargin{1em}

The entire procedure is presented as pseudo-code in Algorithm \ref{alg:selfcal}.
The algorithm shows remarkable convergence properties; it has been observed to
fail only when the sampling of the focal space was not sufficiently dense (in
practice, less than twenty focal values in each direction), and therefore all the
tested infinity planes were not close enough to the correct one. Such problems
are easy to detect, since they usually take the final, refined solution outside
the legal search space.

In principle, autocalibration requires a minimum number of images to work,
according to the autocalibration ``counting argument'' \cite{LuoVie96} (e.g.~$4$
images with known skew and aspect ratio).  However, as we strive to maintain an
``almost'' Euclidean reference frame from the beginning, to better condition
subsequent processing, autocalibration is triggered for models starting from two
images.  The result is an approximate Euclidean upgrade; in fact these models
are still regarded as projective, until they reach a sufficient cardinality.
After that point autocalibration is not performed any more and the internal
parameters of each camera are refined further only with bundle adjustment, as
the computation proceeds. In order not to hamper the process too much, the
internal parameters of a camera becomes fixed after they have been
bundle-adjusted together with a given number of cameras.

\subsection{Projective stereo-modeling.}
\label{sec:2vu}
The model that can be obtained from two uncalibrated images is always
projective.
The following two camera matrices are used:
\begin{equation}
P_1 = [I \ \vert \ \mathbf{0}] \quad \text{and} \quad P_2 = 
[[\mathbf{e}_2]_{\times}F  \ \vert  \  \mathbf{e}_2],
\end{equation}
This canonical pair yields a projective model  with the plane at
infinity passing through the centre of the second camera, which is very unnatural.
Therefore, the method of Section \ref{sec:autocal1} is applied for guessing a
better location for the plane at infinity compatible with rough focal estimates,
obtained from the magnitude of the image diagonal.  Even when the true focal
lengths are far from the estimates, this procedure will provide a useful, well
conditioned starting point for the subsequent steps.

Cheirality is then tested and enforced on the model. In practice only a
reflection (switches all points from in front to behind the cameras) may be
necessary, as the twisted pair case never occurs. In fact, the twisting
corresponds to the infinity plane crossing the baseline, which would imply that
our guess for the infinity plane is indeed very poor.

The 3D coordinates of the tie-points are then obtained by intersection as
before. Finally bundle adjustment is run to improve the model.

\subsection{Resection-intersection.}
\label{sec:res-int}

The procedure is the same as in the calibrated case, taking care of using the
Direct Linear Transform (DLT) algorithm \cite{HarZis00} for resection, as the the
single image is always uncalibrated. While PPnP computes only the external
parameters of the camera, the DLT computes the full camera matrix.

\subsection{Merging two models.}
\label{sec:merge}

Partial models live in two different reference frames, that are related by a
similarity – if both are Euclidean – or by a projectivity – if one is
projective.
In this case the projectivity that brings the projective model onto the
Euclidean one is sought, thereby recovering its correct Euclidean reference frame.
The procedure is the same as in the calibrated case, with the only difference
that when computing the projectivity the DLT algorithm should be used instead of
OP.

The new model is refined with bundle adjustment (either Euclidean or projective)
and upgraded to a Euclidean frame when the conditions stated beforehand are met.

\section{Parameter Settings}
\label{sec:devil} 

\sammy is a complex pipeline with many internal parameters. With respect to this
issue our endeavor was: i) to avoid free parameters at all; ii) to make
them data-dependent; iii) to make user-specified parameters intelligible and
subject to an educated guess.  In the last case a default should be provided
that works with most scenarios. This guarantees the processing to be
completely automatic in the majority of cases.

All the heuristic parameter settings used in the experiments have been reported
and summarized in Table \ref{tab:parameters}.

\begin{table}[htbp]
\centering
\begin{tabular}{| l | c |}
  \hline    
  \textbf{Parameter}   & \textbf{Value} \\
  \hline    
  \multicolumn{2}{| c |}{ \it{Keypoint detection} }  \\
  \hline    
  Number of LoG pyramid levels  & 12 \\
  Average number of keypoints per image  & 7500 \\
  \hline    
  \multicolumn{2}{| c |}{ \it{Matching - broad} }  \\
  \hline    
  Number of keypoints per image   & 300 \\  
  Degree of edge-connectedness & 8 \\ 
 \hline    
  \multicolumn{2}{| c |}{ \it{Matching - narrow} }  \\
  \hline   
  Matching discriminative ratio  & 1.5 \\
  Maximum number of MSAC iterations  & 1000 \\
  Bucket size  (MSAC) in pixels & $ D/25 $ \\
  Minimum number of  matches  & 10 \\ 
  Homography-to-Fundamental GRIC Ratio & 1.2 \\
  Minimum track length  & 3 \\
\hline
 \multicolumn{2}{| c |}{ \it{Reconstruction } }  \\
  \hline    
  Maximum bundle adjustment iterations  & 100 \\
  Reprojection Error  & $ D/1800 $ \\
  \hline  
  \multicolumn{2}{| c |}{ \it{Autocalibration} }  \\
  \hline    
  Number of cameras for  autocalibration  & 4 \\ 
  Number of cameras  to fix internal param.s  & 25 \\
   \hline    
  \multicolumn{2}{| c |}{ \it{Prologue} }  \\
  \hline    
  Final minimum track length  & 2 \\
  Final maximum reprojection error  & $D/2400 $ \\
  \hline    
\end{tabular}
\smallskip 
\caption{Main settable parameters of \sammy. $D$ is the image diagonal length [pixels].}
\label{tab:parameters}
\end{table}

\paragraph{Keypoint detection}

 The keypoints extracted from all images are ordered by their response value and
 the ones with the highest response are retained, while the others are
 discarded. The total number of keypoints to be kept is a multiple of the number
 of images, so as to keep the average quota of keypoints for each image fixed.

\paragraph{Matching -  broad phase}

During the broad matching phase the goal is to compute the 2D histogram
mentioned in Sec.~\ref{sec:broad}. To this end, the keypoints in each image are
ordered by scale and the 300 keypoints with the highest scale are considered.
The number of neighbors in feature space is set to six, as in
\cite{Brolow03}. Given the nature of the broad phase, this value is not
critical, to the point where only the number of keypoints is exposed as a
parameter.

The number $m$ in Sec.~\ref{sec:broad} (``Degree of edge-connected\-ness'', in
Table \ref{tab:parameters}), has been set to eight following \cite{Brolow03}. In
our case, the role of the parameter is more ``global'', as it does not set
 the exact number of images to be matched but the degree of edge-connectedness
of the graph. However our experiments confirmed that that $m=8$ is a good default.

\paragraph{Matching - narrow phase}

 The number of max iterations of MSAC is set to $1000$ during the matching
 phase. This is only an upper bound, for the actual value is dynamically updated
 every time a new set of inliers is found.

The ``Matching discriminative ratio'' refers to the ratio of first to second
closest keypoint descriptors, used to prune weak matches at the beginning of this
phase.

The `` Minimum number of matches'' parameter refers to the last stage of the
narrow phase, when poor matches between images are discarded based on the number
of surviving inliers after MSAC.

\begin{figure}[t]
  \centering
  \includegraphics[width=1\linewidth]{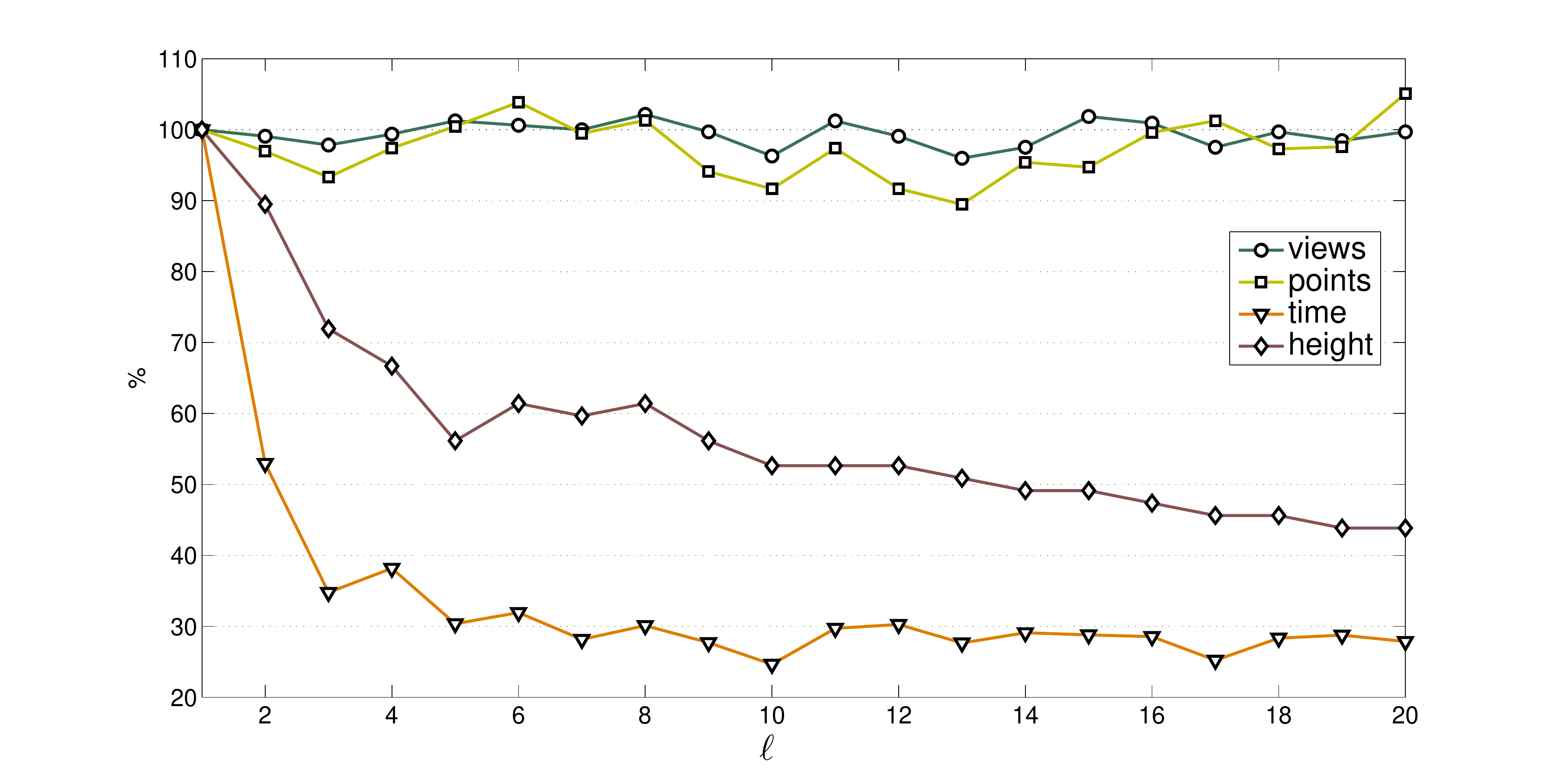}
  \caption{This plot shows the number of reconstructed tie-points, images, height of
    the tree and computing time as a function of the parameter $\ell$ in the
    balancing heuristics. The values on the ordinate are in percentage,≈ß where
     the baseline case $\ell=1$.}\label{fig:bilancia}
 \end{figure}

\paragraph{Clustering}
The parameter $\ell$ of Sec.~\ref{sec:balance} has been set to $\ell = 3$ based
on the graph reported in Fig.~\ref{fig:bilancia}, where the
number of reconstructed tie-points/images and the computing time are plotted as
the value of $\ell$ is increased. After $\ell = 3$, the computing time
stabilizes at around 30\% of the baseline case, without any significant
difference in terms of number of reconstructed images and tie-points.

\paragraph{Reconstruction}
The safeguard threshold on the reprojection error (Sec.~\ref{sec:inter}) is set
to 2 pixels with reference to a 6 Mpixel image, and scaled according to the
actual image diagonal (assuming an aspect ratio of 4:3, the reference diagonal
is $3600$ pixels).

\paragraph{Autocalibration}

The Euclidean upgrade is stopped as soon as a cluster reaches a sufficient
cardinality $k$ (``Number of cameras for autocalibration'') that satisfies the
following inequality \cite{LuoVie96}, giving the condition under which
autocalibration is feasible:
\begin{equation}
5k - 8 \geq (k-1) (5 - p_k - p_c) + 5 - p_k
\end{equation}
where $p_k$ internal parameters are known and $p_c$ internal parameters are
constant.  With known (or guessed) skew and aspect ratio ($p_k=2$ and $p_c=0$)
four cameras are sufficient. 

The reason for keeping this value  to the minimum
is because we observed experimentally that projective alignment is fairly
unstable and it is beneficial to start using a similarity transformation as soon as
possible.

The cluster cardinality after which the internal parameters are kept fixed in
the bundle adjustment is set to 25, a fairly high value, that guarantees all the
internals parameters, especially the radial distortion ones, are steady.

\paragraph{Local bundle adjustment} As discussed previously, the local bundle 
adjustment is generally to be preferred over the full one. However, since the
autocalibration phase is crucial, our strategy is to run the full bundle
adjustment  until the clusters become Euclidean. It should also be noted that the
computational benefits of the local bundle adjustment are more evident with
large clusters of cameras.

\paragraph{Prologue}
 The last bundle adjustment is always full (not local) and is run with a lower
 safeguard threshold on the reprojection error (1.5 pixel in the reference 6
 Mpixel image). A final intersection step is carried out using also the tracks
 of length two, in order to increase the density of the model. Please note
 however that these weak points do not interfere with the bundle adjustment, as
 they are added only \emph{after} it.

\section{Experiments \label{sec:exp}}

We run \sammy on several real, challenging datasets, summarized in
Tab.~\ref{tab:summary}. All of them have some ground truth available, being
either a point cloud (from laser scanning), the camera internal parameters or
measured ``ground'' control points (GCP).

The qualitative comparison against \vsfm focuses on differences in terms of
number of reconstructed cameras and manifest errors in the model.  The
quantitative evaluation examines the accuracy of the model against GCPs and
laser scans, and/or the internal orientation accuracy.

\begin{table*}[htbp]
\centering
\caption{Summary of the experiments. The ``camera'' column refers to the number of
  different cameras or to different settings of the same camera.}
\smallskip 
\begin{tabular}{l r c l l l}
  \hline        & \it \#images    & \it \#cameras  & \it ground truth     & \it notes \\[2pt] 
   Br\`a   & 331  & 1    & laser              &   laser/photo inconsistencies \\ 
  Duomo di Pisa  & 309  & 3    & laser        &    3 cameras  \\   
  S.~Giacomo     & 270  & 2    & laser          &  2 sets 1 year apart\\  
  Navona         & 92   & 1    & internal     &    also used in \cite{remondino2012low}\\  
  Ventimiglia    & 477  & 1   & GCP              &  nadiral/oblique \\   
  Termica        & 27   & 1    & GCP              & nadiral \\            
  Herz-Jesu-P25  & 25   & 1    & GCP, internal  &  reference dataset \cite{strecha2008} \\    
\hline 
\end{tabular}
\label{tab:summary}
\end{table*}

\subsection{Qualitative evaluation.}

We compared \sammy (version 1.3) with \vsfm (version 0.5.22)
\cite{VisualSfm,MulticoreBA}, a recent sequential pipeline that improves on
Bundler \cite{SnaSeiSze06} in terms of robustness and efficiency.

In all the experiments of this section, the internal camera parameters were
considered to be unknown for both pipelines. However, while \sammy uses a
completely autocalibrated approach, \vsfm extracts an initial guess of the
internal parameters from the image EXIF and a database of known sensor sizes for
common cameras (or uses a default guess for the focal length).

\begin{figure*}[htbp]
\centerline {
\fbox{\includegraphics[width=0.49\textwidth]{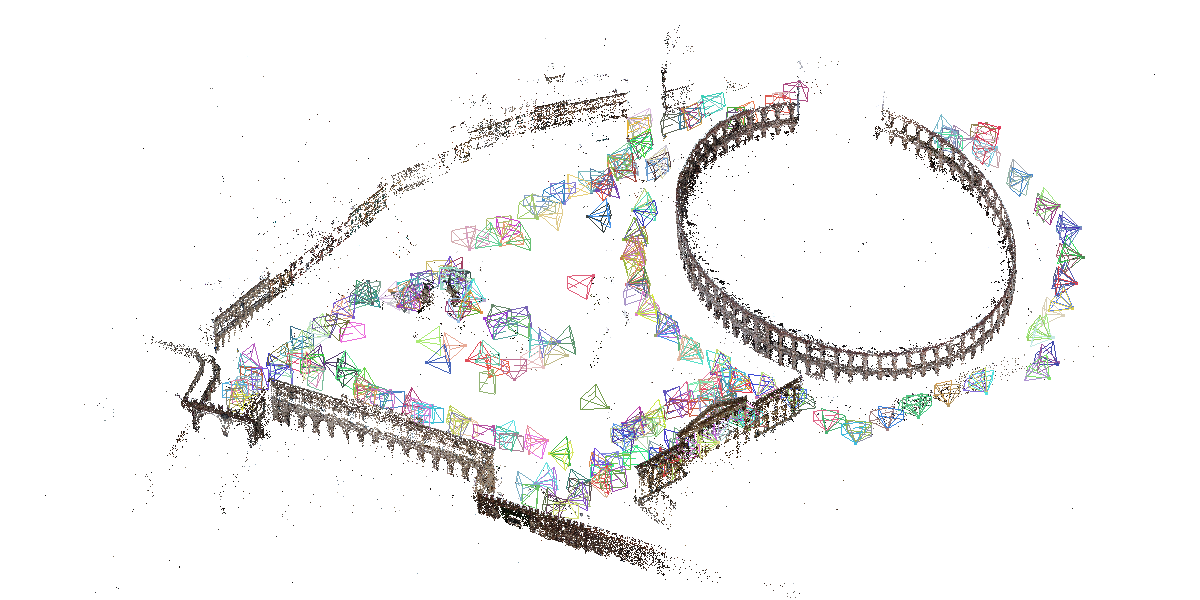}} \hfill
\fbox{\includegraphics[width=0.49\textwidth]{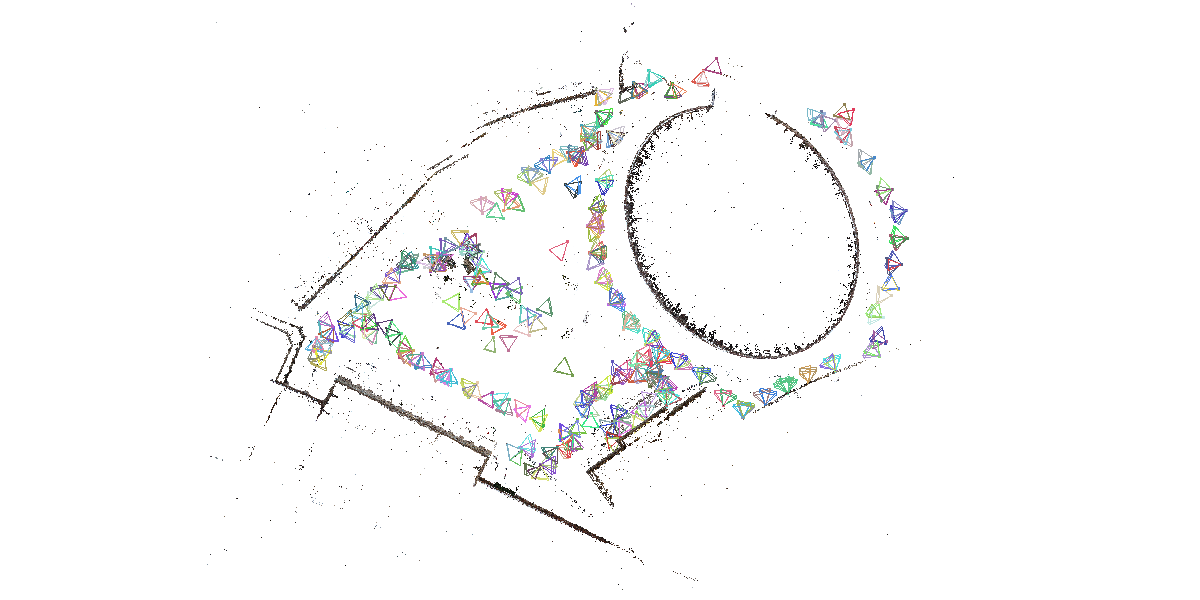}}
}
\smallskip
\centerline {
\fbox{\includegraphics[width=0.49\textwidth]{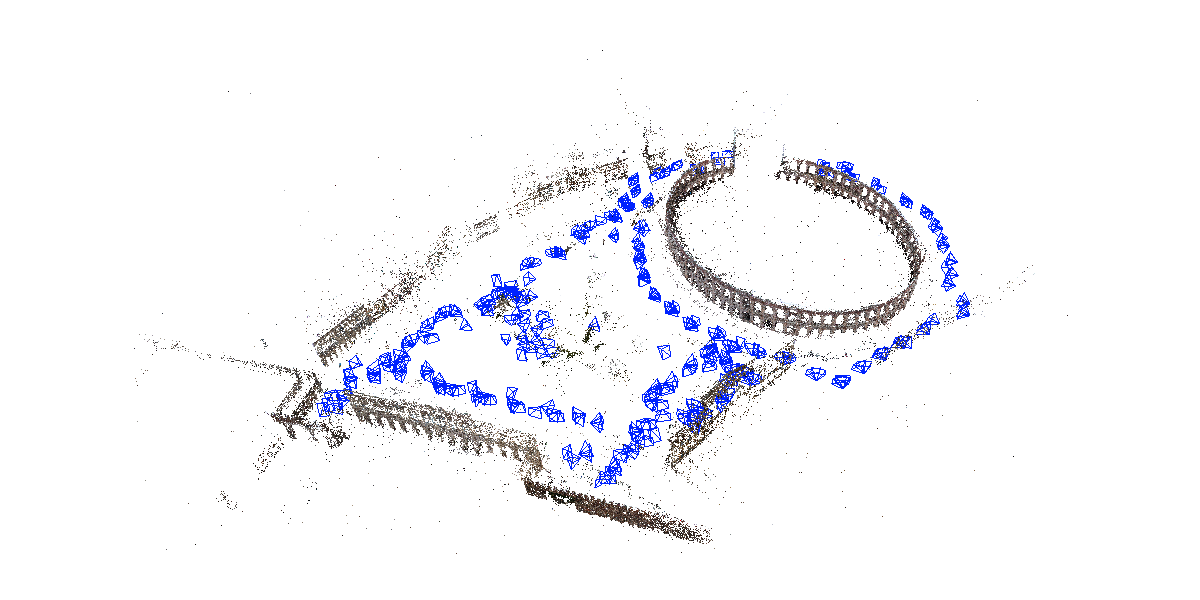}} \hfill
\fbox{\includegraphics[width=0.49\textwidth]{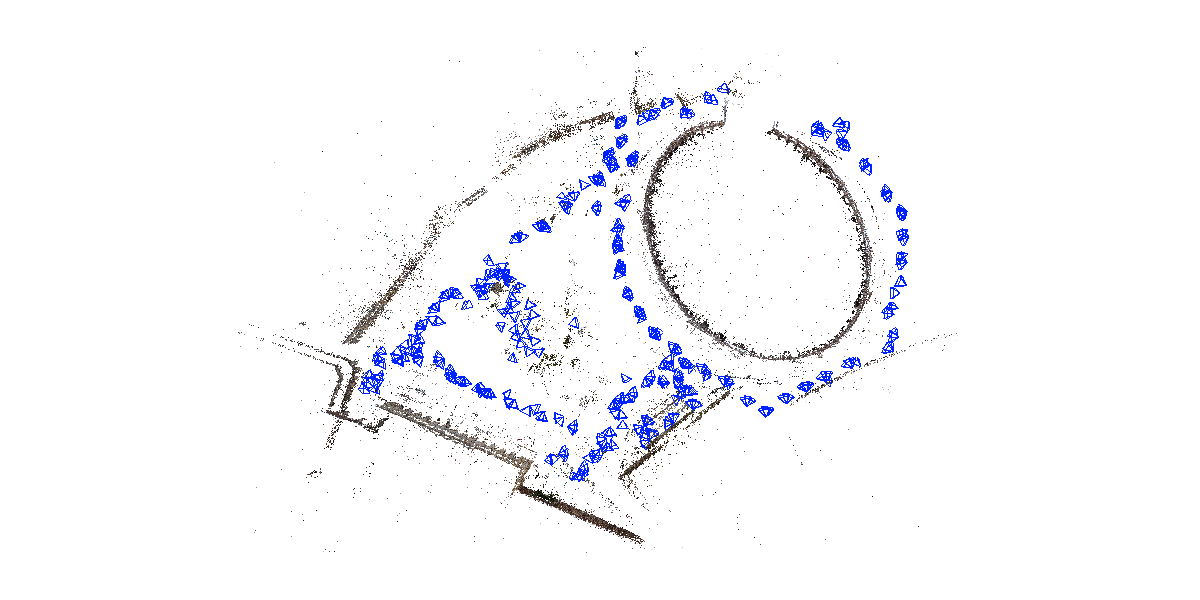}}
}
\caption{Comparative modeling results from the ``Br\`a'' dataset. Top:
  \vsfm. Bottom: \sammy.  In this case both methods produced
    visually correct results.}
\label{ResultsSamantha_Bra}
\end{figure*}

We report four experiments.  The first one, ``Br\`a'', is composed of 331
photos of Piazza Br\`a, the main square in Verona where the Arena is
located. Photos were taken with a Nikon D50 camera and a fixed focal length of
18mm. This dataset is also  publicly available for
download \footnote{http://www.diegm.uniud.it/fusiello/demo/samantha/}. Results
are shown in Figure \ref{ResultsSamantha_Bra}. Both \vsfm and \sammy produced a
correct result in this case. It can also be noticed that Bundler failed 
with the same dataset \cite{GheFarFus10}.

\begin{figure*}[htbp]
\centerline {
  \fbox{\includegraphics[width=0.49\textwidth]{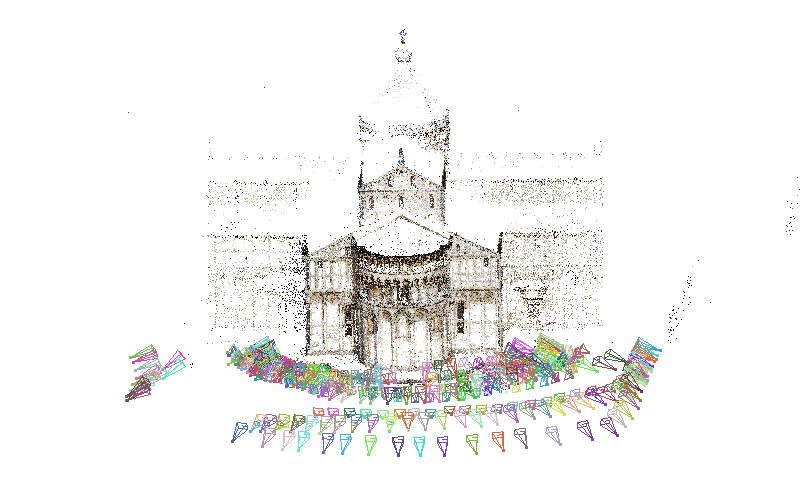}}
  \hfill
  \fbox{\includegraphics[width=0.49\textwidth]{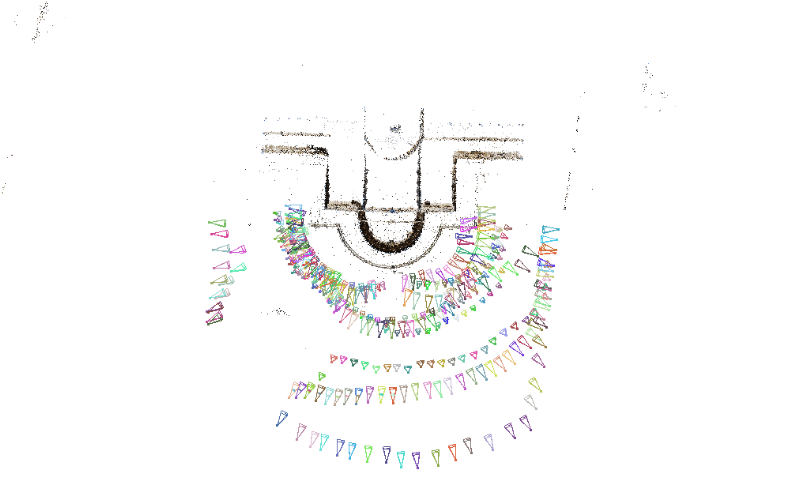}} }
\smallskip
\centerline {
\fbox{\includegraphics[width=0.49\textwidth]{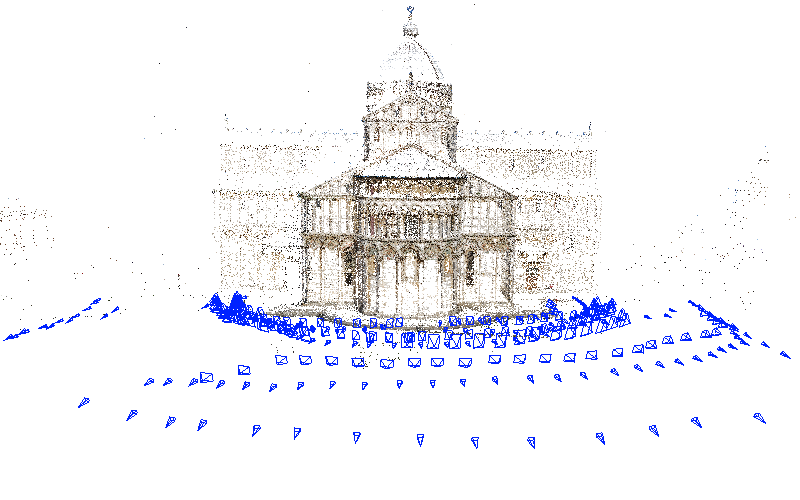}} \hfill
\fbox{\includegraphics[width=0.49\textwidth]{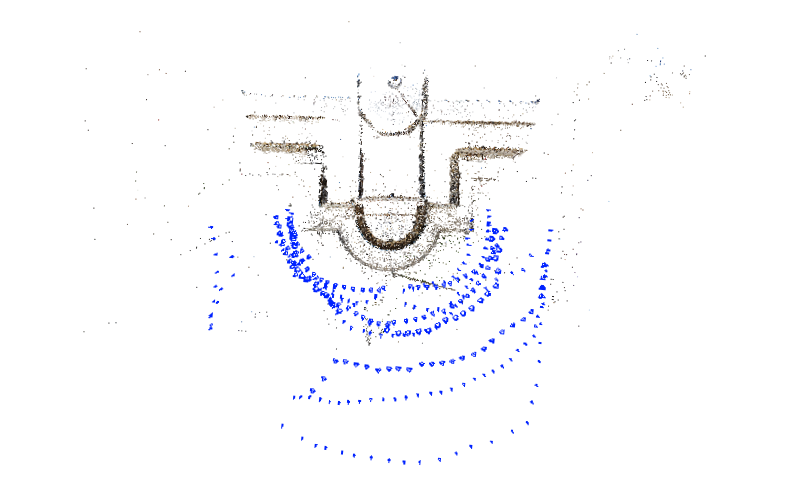}}
}
\caption{Comparative modeling results from the  ``Duomo di Pisa'' dataset. Top:
  \vsfm. Bottom: \sammy.  In this case both methods produced
    visually correct results.}
\label{ResultsSamantha_Duomo}
\end{figure*}

The second test, ``Duomo di Pisa'', is composed of 309 photos of the Duomo of
Pisa in the famous Piazza dei Miracoli square. The dataset is composed of three
sets of photos taken with a Nikon D40X camera at different focal lengths (13mm,
20mm and 58mm).  Results are shown in Figure \ref{ResultsSamantha_Duomo}. Also
in this case both the pipelines produced a good solution. This dataset is a
relatively easy one as far as the \sam is concerned, for there are many photos
covering a relatively small and limited scene, but there are three different
cameras, which makes it challenging for the autocalibration.

\begin{figure*}[htbp]
\centerline {
\fbox{\includegraphics[width=0.49\textwidth]{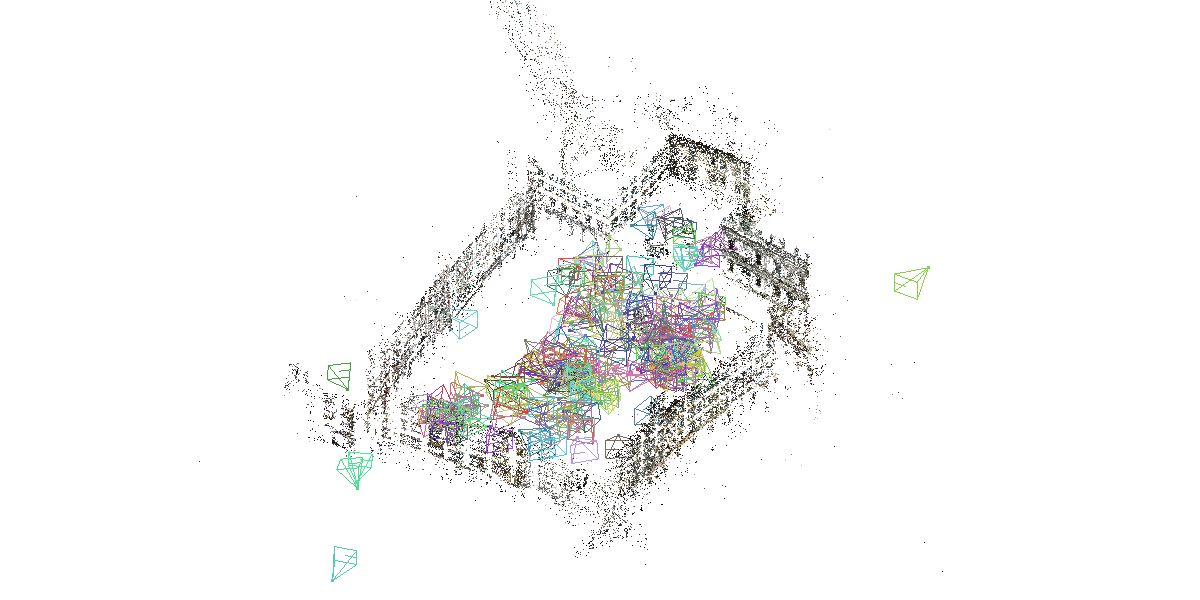}} \hfill
\fbox{\includegraphics[width=0.49\textwidth]{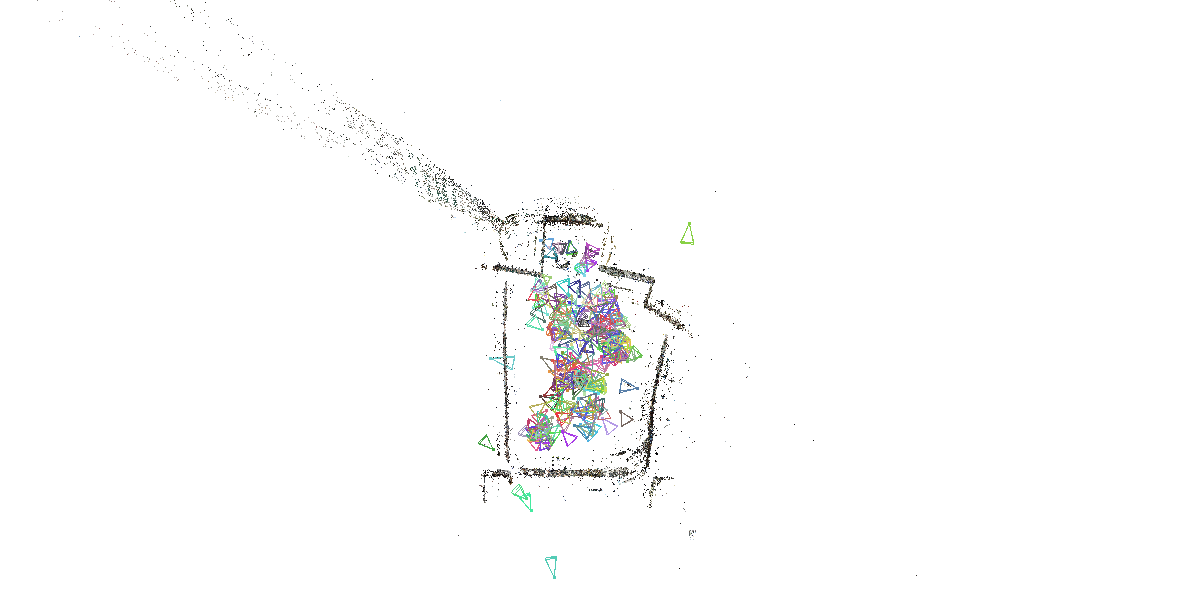}}
}
\smallskip
\centerline {
\fbox{\includegraphics[width=0.49\textwidth]{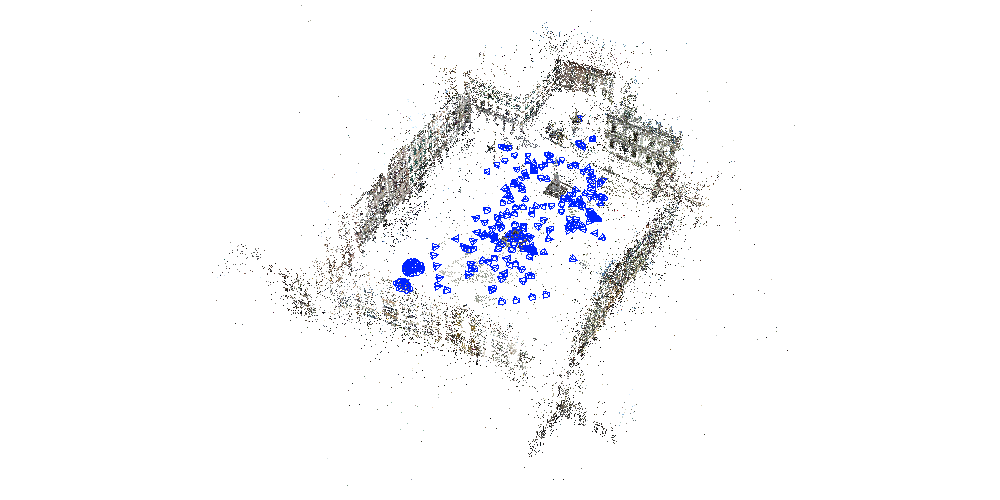}} \hfill
\fbox{\includegraphics[width=0.49\textwidth]{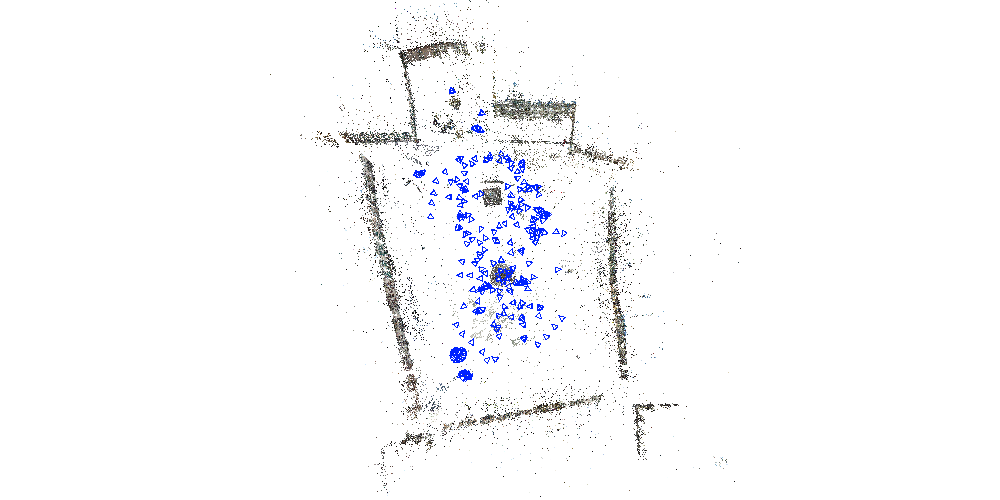}}
}
\caption{Comparative modeling results from the ``S. Giacomo''
  dataset. Top: \vsfm. Bottom: \sammy.
  Please note the cameras outside the square and the rogue points in the
  \vsfm model.
 \label{ResultsSamantha_ErbeUdine} }
\end{figure*}

The third tested dataset, ``S. Giacomo'', is composed of 270 photos of one of
the main squares in Udine (IT). It consists of two set of exposures. The first
one was shot with a Sony DSC-H10 camera with a fixed focal length of 6mm
while the second one -- one year after the first -- with a Pentax OptioE20
camera with a fixed focal length of 6 mm. Results are shown in Figure
\ref{ResultsSamantha_ErbeUdine}.  While \sammy produced a visually
  correct result, some cameras of \vsfm are located out of the square and some
  walls are manifestly wrong.

\begin{figure*}[htbp]
\centerline {ù
\fbox{\includegraphics[width=0.49\textwidth]{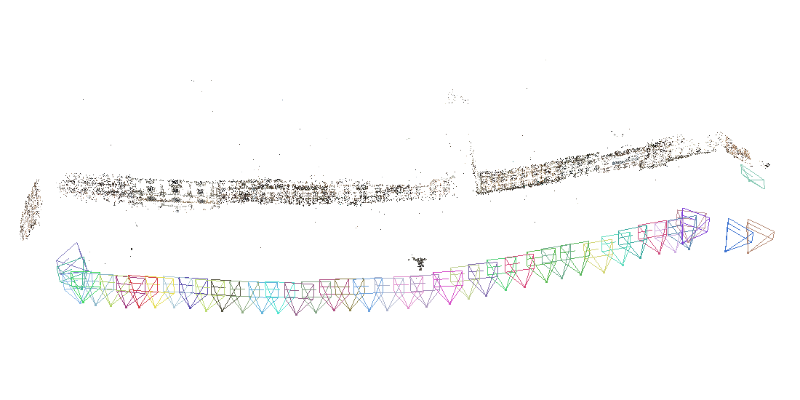}} \hfill
\fbox{\includegraphics[width=0.49\textwidth]{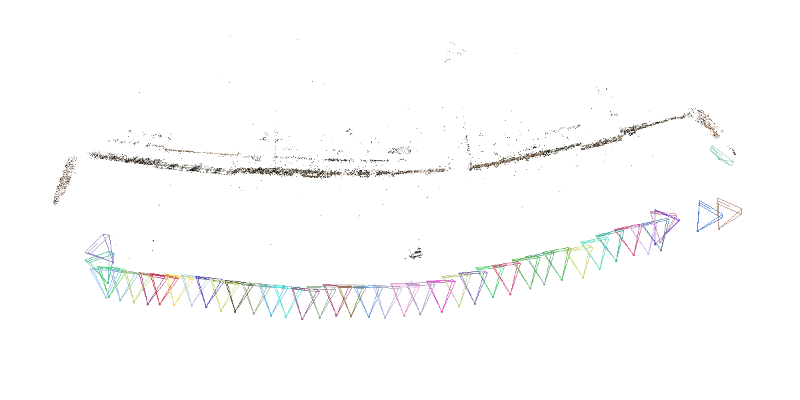}}
}
\smallskip
\centerline {
\fbox{\includegraphics[width=0.49\textwidth]{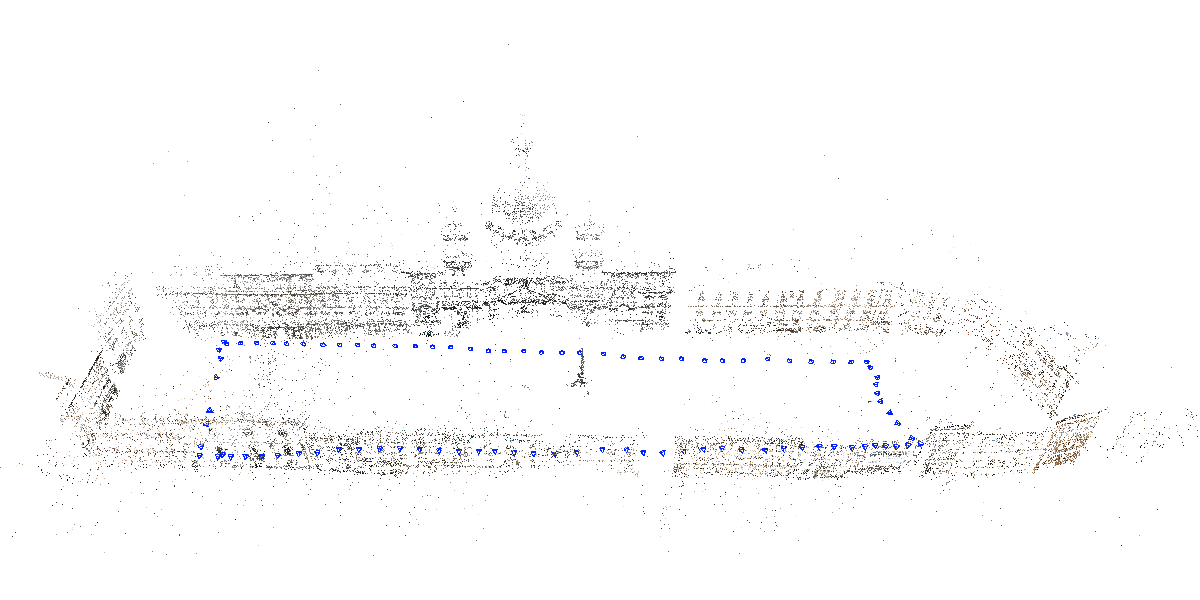}} \hfill
\fbox{\includegraphics[width=0.49\textwidth]{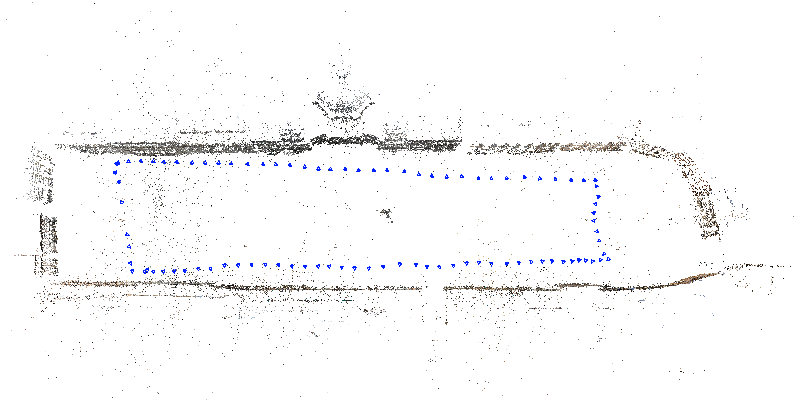}}
}
\caption{Comparative modeling results from the ``Navona'' dataset. Top:
  \vsfm. Bottom: \sammy.  Please note that 
  \vsfm modeled only half of the square and  the facade is bent.} 
\label{ResultsSamantha_Navona}
\end{figure*}

The last set, ``Navona'', contains 92 photos of the famous square in Rome, taken
with a Samsung ST45 Camera at a fixed focal length of 35mm.  The dataset is
publicly available for
download\footnote{http://www.icet-rilevamento.lecco.polimi.it/}.
Results are shown in Figure \ref{ResultsSamantha_Navona}. In this case, \sammy
produced a complete and correct model of the square while \vsfm produced a
partial and incorrect model, as reported also in \cite{remondino2012low}.

For this dataset the internal parameters of the camera were also available (from
off-line standard calibration), thus allowing us to compare the focal length
obtained by autocalibration, which achieved an error of 2.3\%
(Tab.~\ref{tab:comparisonK}).

\begin{table}[tb]
\centering
\caption{Comparison with ground control points (GCP). All measures are in mm.}
\smallskip
\begin{tabular}{lrrr}
\hline 
               &  RMS error & avg. depth &  GSD   \\
Ventimiglia        & 22~  &  15 $\cdot 10^3$~ &  $<$ 4  \\
Termica            & 86~   &  79 $\cdot 10^3$~ & 27   \\  
Herz-Jesu-P25      & 3.4   & 3.4 $\cdot 10^3$ &  5    \\  
\hline
\end{tabular}
\label{tab:comparisonPPM}
\end{table}

\subsection{Comparison against control points.}

In this set of experiments we tested the models obtained by \sammy in a context
where the position of some ``ground'' control points (GCP) was  measured
independently (by GPS or other techniques).  These control points was
identified manually in the images and their position in space was
estimated by intersection.  Correspondences between true and estimated 3D
coordinates have been used to transform the model with a similarity that aligns
the control points in the least-squares sense. The root mean square (RMS)
residual of this registration was taken as an indicator of the accuracy of
the model.

\begin{figure*}[htbp]
\centerline {
  \fbox{\includegraphics[width=0.32\textwidth]{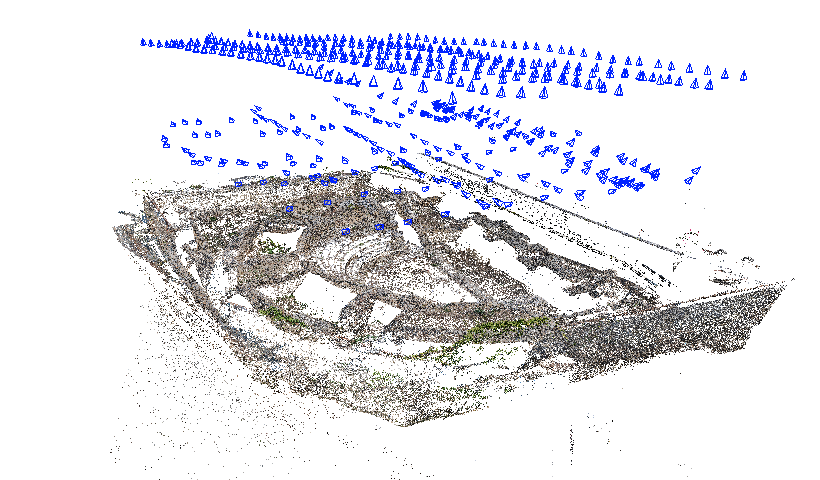}}
  \hfill
  \fbox{\includegraphics[width=0.32\textwidth]{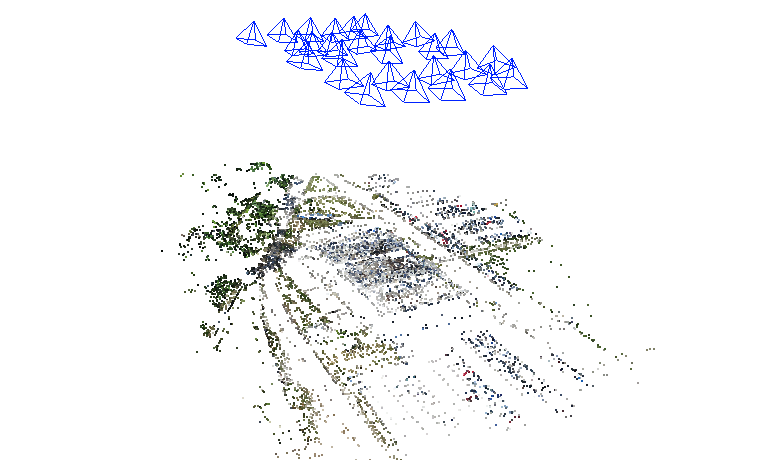}}
  \hfill
  \fbox{\includegraphics[width=0.32\textwidth]{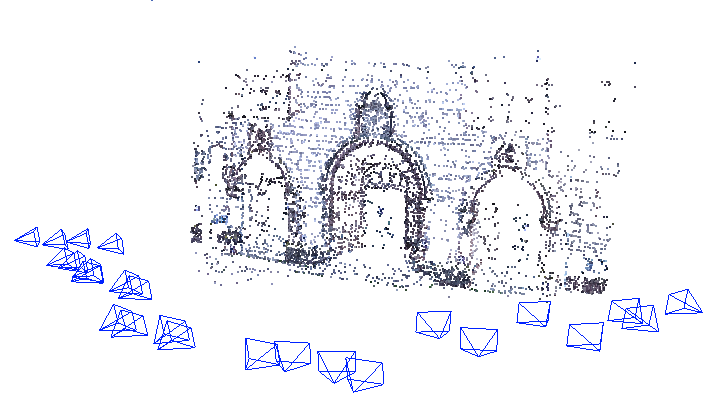}} }
\caption{From left to right: results for ``Ventimiglia'', ``Termica'',
  ``Herz-Jesu-P25'' obtained with \sammy}
\label{fig:3wgcp}
\end{figure*}

\begin{figure*}[htbp]
\centering
\begin{minipage}{0.48\linewidth}
\centering
\centering 
\includegraphics[width=0.999\textwidth]{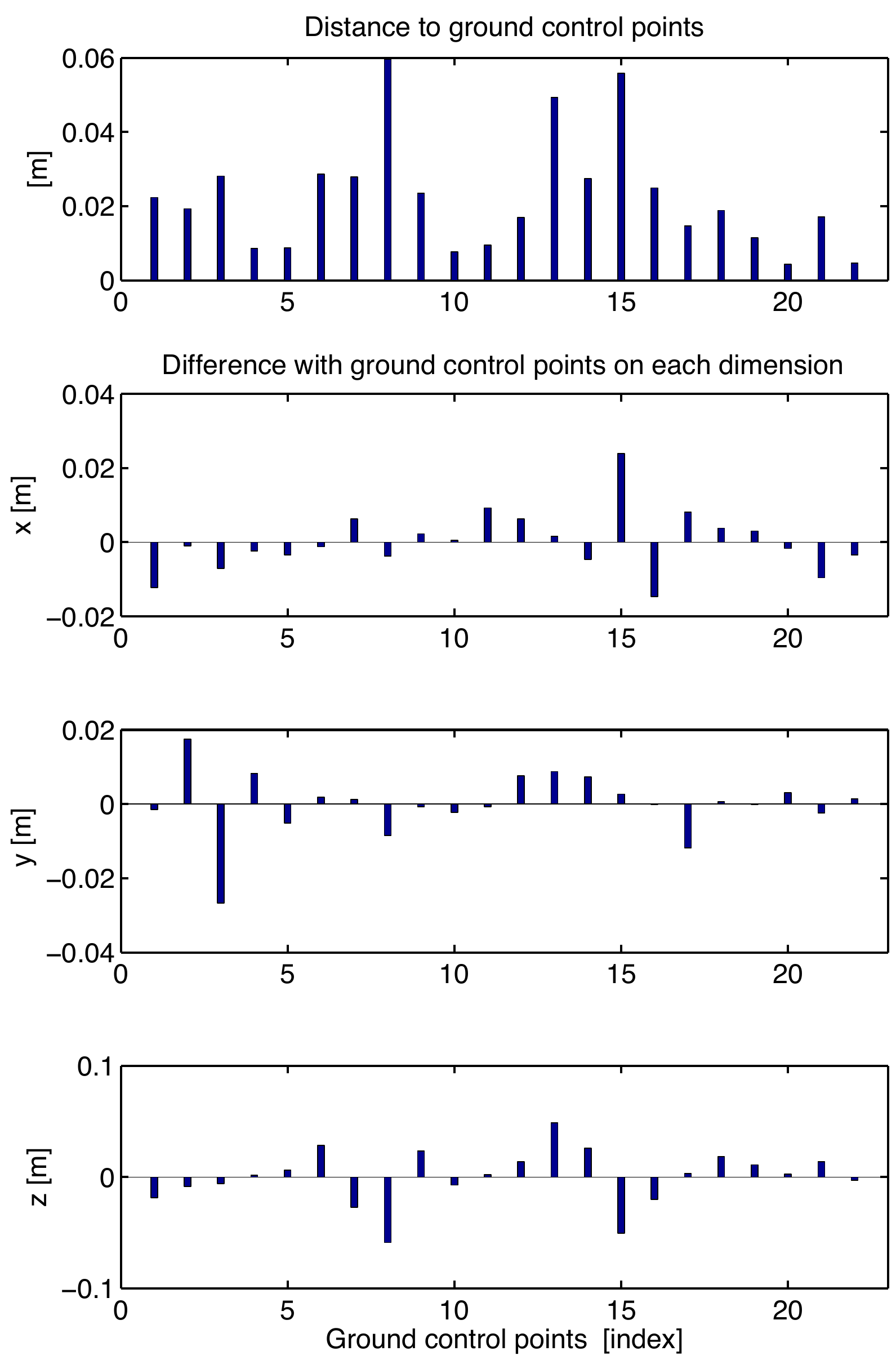}
\caption{Top: Distance of computed 3D points to ground control points for the
  ``Ventimiglia'' image-set. Bottom: Differences on each dimension
  (X-Y-Z). }\label{fig:ventimiglia_dist}
\medskip
\end{minipage}
\hfill
\begin{minipage}{0.48\linewidth}
\centering
\centering 
\includegraphics[width=0.999\textwidth]{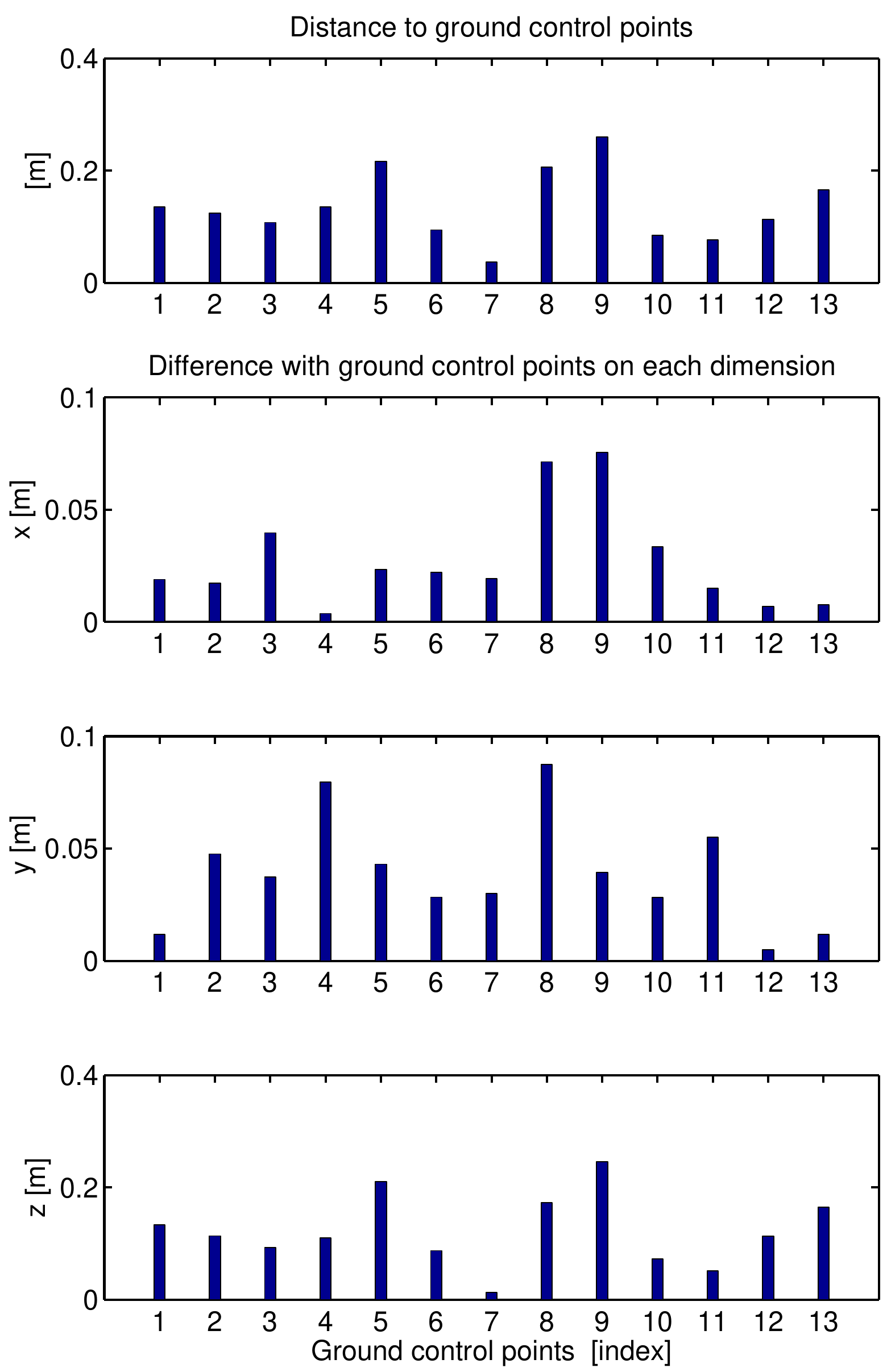}
\caption{Top: Distance of computed 3D points to ground control points for the
  ``Termica''  image-set. Bottom: Differences on each dimension
  (X-Y-Z). }\label{fig:termica_dist}
\medskip
\end{minipage}
\end{figure*}

``Ventimiglia'' (477 photos) is composed of aerial photos -- nadiral and oblique
-- taken with a RPA\footnote{Remotely Piloted Aircraft} equipped with a 24 Mpixel Nikon D3X (full frame sensor) mounting a Nikkon 50 mm lens.  23
targeted points have been measured through topographic surveying using a Topcon
GPT-7001i total station. The (average) ground sampling distance (GSD) is less
than 4 mm.  After modeling with \sammy and least-squares alignment, the RMS
error with respect to the control points is 22 mm. Figure
\ref{fig:ventimiglia_dist} reports  the individual errors.

``Termica'' (27 images) was captured with a RPA equipped with a 12 Mpixel
Canon Power Shot S100 camera (1/1.7’’ CMOS sensor) with nadiral attitude.  13
non-signalized natural target points was measured by geodetic-grade GPS
receivers.  The (average) GSD is 27 mm. After modeling with \sammy and
least-squares alignment, the RMS error with respect to the control points is 86
mm,  which is well within the uncertainty affecting the measured
  camera positions (reported in \cite{strecha2008}). Individual errors are
shown in Figure \ref{fig:termica_dist}.

\begin{figure}[htbp]
\centering 
\includegraphics[width=0.48\textwidth]{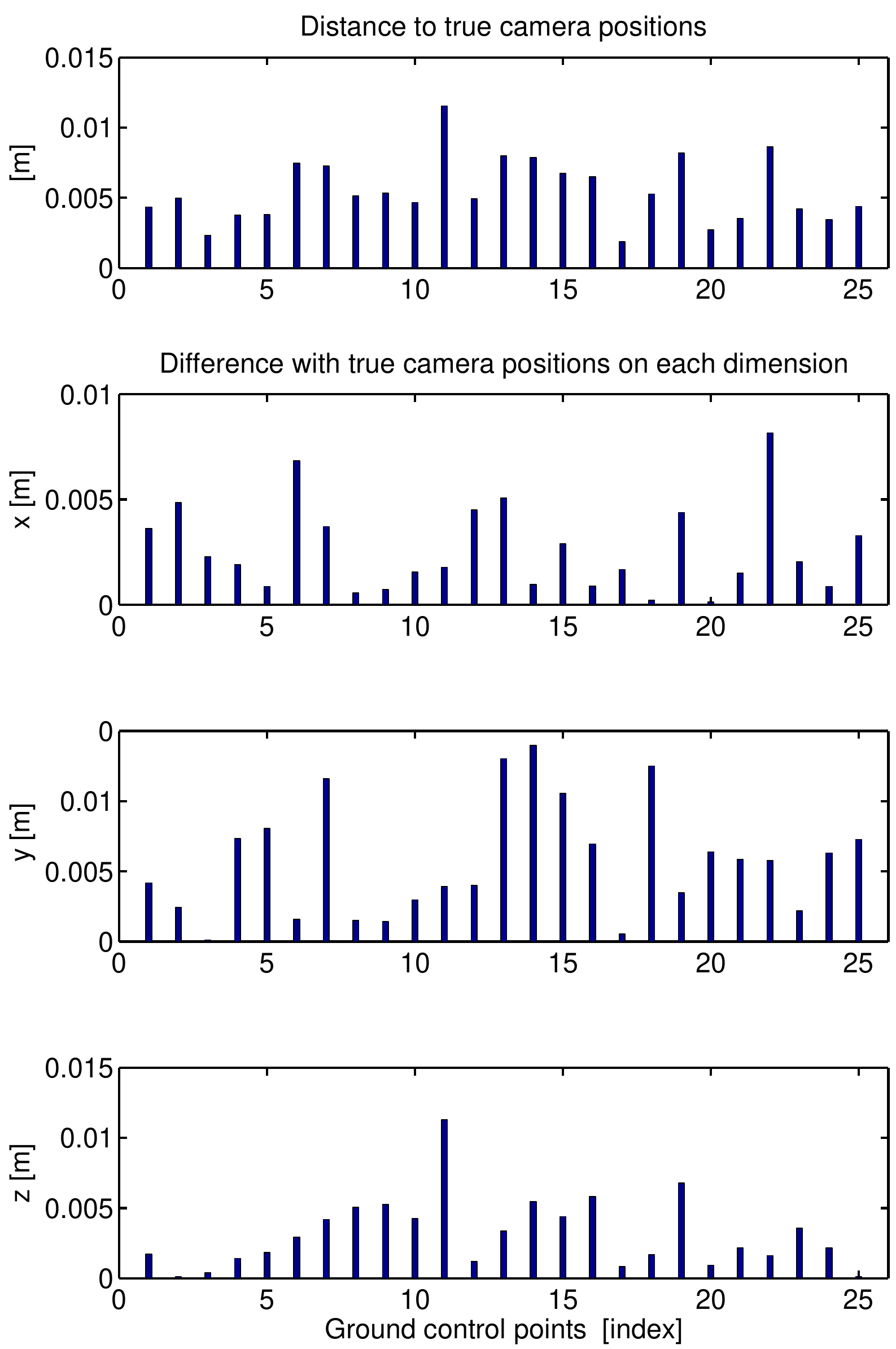}
\caption{Top: Distance of computed camera centers to reference positions for the
  ``Herz-Jesu-P25'' image-set. Bottom: Differences on each dimension
  (X-Y-Z). }\label{fig:hz25_dist}
\end{figure}

``Herz-Jesu-P25'' is part of a publicly available dataset \cite{strecha2008}, it
consist of 25 cameras, for which the ground truth position and attitude had been
computed via alignment with a laser model; in this case, as GCPs we considered
the camera centres.
After modeling with \sammy and least-squares alignment, the RMS error
with respect to the control points is 3.4 mm.
Figure \ref{fig:hz25_dist}  reports the individual errors.

For this dataset  the internal parameters of the camera were also available, thus
allowing us to validate the parameters obtained by autocalibration, with a
remarkably low error of 0.2\% on the focal length value
(Tab.~\ref{tab:comparisonK}).

\begin{figure*}[htbp]
\centerline {
\fbox{\includegraphics[width=0.32\textwidth]{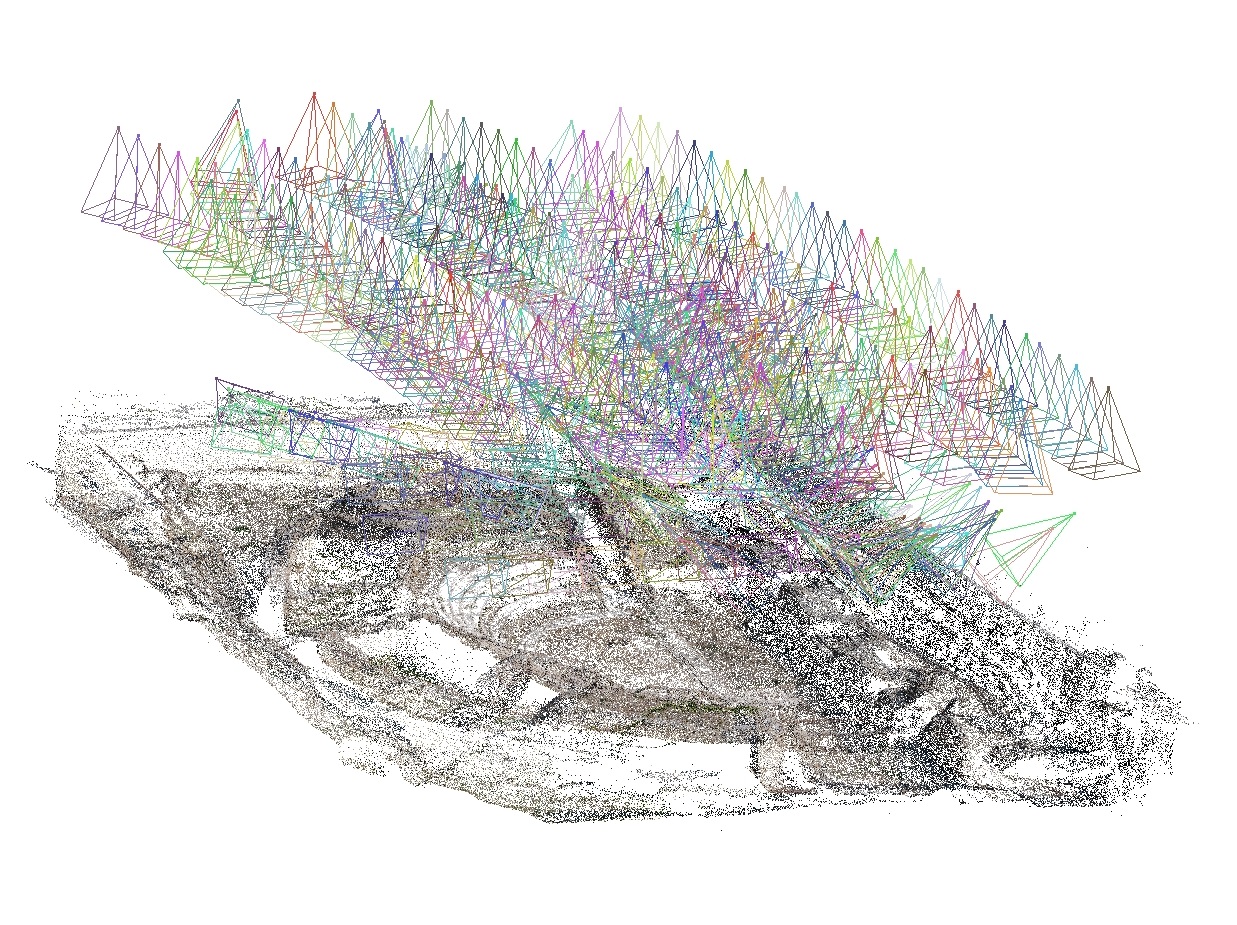}} \hfill
\fbox{\includegraphics[width=0.32\textwidth]{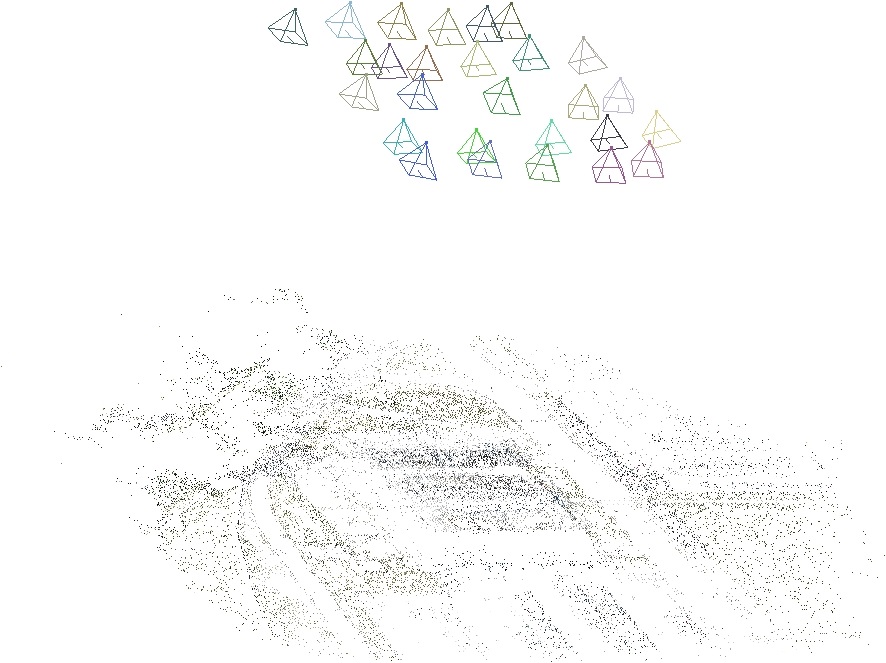}}  \hfill
\fbox{\includegraphics[width=0.32\textwidth]{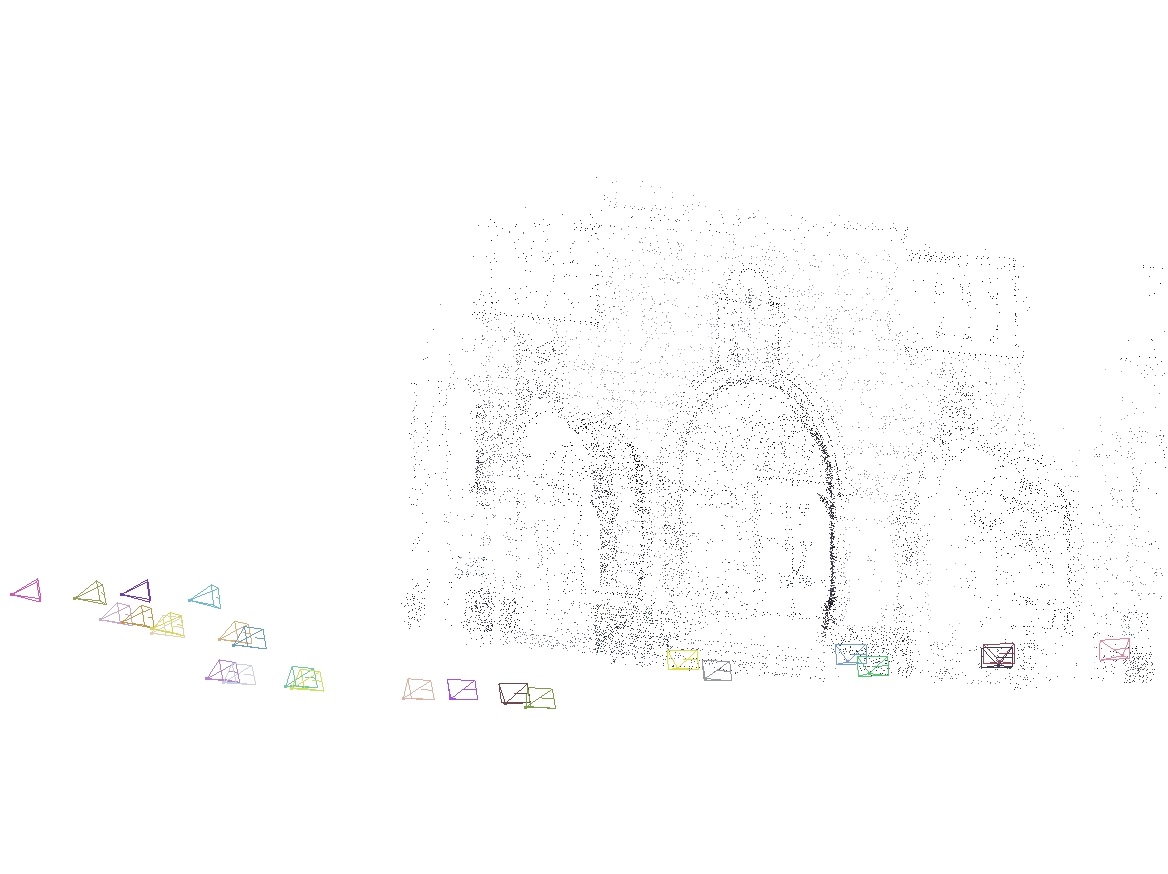}}
}
\caption{From left to right: results for ``Ventimiglia'', ``Termica'',
  ``Herz-Jesu-P25'' obtained with \vsfm}
\label{fig:3wgcpVsfm}
\end{figure*}

Figures \ref{fig:3wgcp} shows the models obtained for these three datasets.
As a further qualitative comparison, Figures \ref{fig:3wgcpVsfm} shows the
models obtained by \vsfm for the same datasets. Both pipelines produced
qualitatively correct results, but \vsfm discarded a group of 24 photos in
the ``Ventimiglia'' dataset.

\begin{table}[tb]
\centering
\caption{Focal length values (in pixel) and errors [percentage]. The value for
  \vsfm on ``Navona'' missing due to failure. }
\smallskip
\begin{tabular}{llll}
\hline 
&  Calib. & \sammy  & \vsfm   \\
Navona         & 4307.5   & 4208.7  [2.3\%] & n.a. \\  
Herz-Jesu-P25  & 2761.8  & 2756.2  [0.20\%] & 2752.1 [0.32\%] \\ 
\hline
\end{tabular}
\label{tab:comparisonK}
\end{table}

\subsection{Comparison against laser data}

Thanks to the availability of a laser survey for two datasets (``Br\`{a}'' and
``Duomo di Pisa'') we have been able to further assess the accuracy of our
results, by aligning the 3D point produced by \sammy with the laser point cloud
using iterative closest point (ICP) and looking at the residuals (using
``CloudCompare'' software \cite{CloudCompare}).

\begin{figure*}[thbp]
\centerline {
  \fbox{\includegraphics[width=0.49\textwidth]{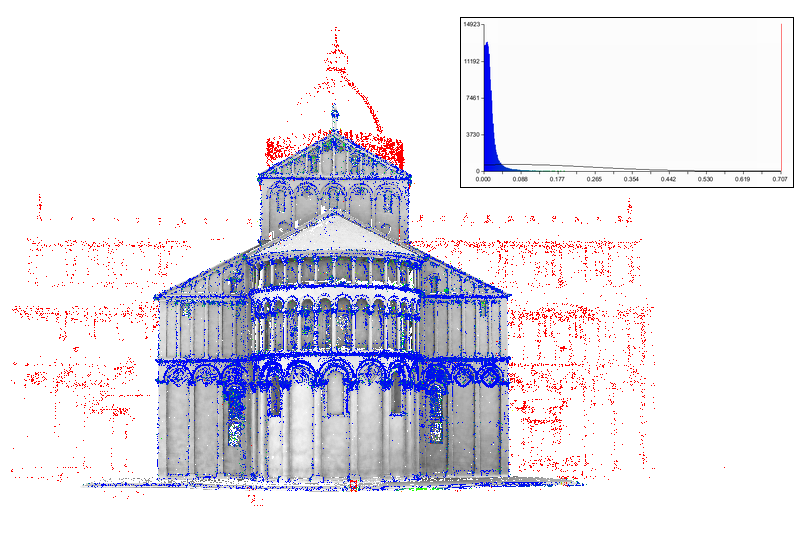}}
  \hfill
  \fbox{\includegraphics[width=0.49\textwidth]{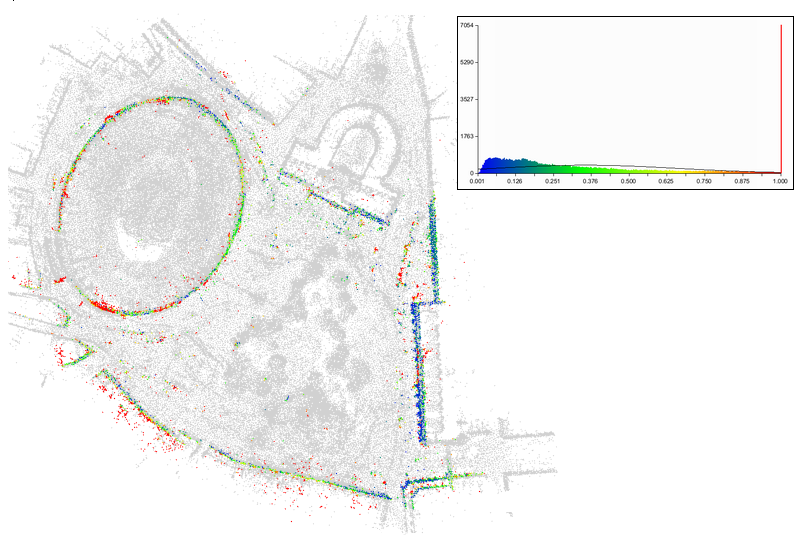}} }
 \caption{Results of point cloud comparison between laser and \sammy for
   ``Br\`{a}'' and ``Duomo di Pisa''. The colour of the \sammy's point cloud
   encodes the residual distance, consistently with the histogram shown in the
   insert (this figure is best viewed in colour). Units are in meters.}\label{fig:comparisonLaser}
\end{figure*}

Results are reported in Tab.~\ref{tab:comparisonLaser} and
Fig.\ref{fig:comparisonLaser} (for ``Duomo di Pisa'' and ``Br\`{a}'' only).

The laser data relative to the ``Duomo di Pisa'' is a triangular mesh with an
average resolution of 30 mm representing a subset of the area surveyed by the
photograph (the abside). After registering the point cloud produced by \sammy
onto this mesh, the residual average point-mesh distance is 73 mm.

The laser data for ``S.~Giacomo'' is a point cloud resampled on a 20 mm
grid. After registering the point cloud produced by \sammy onto this mesh, the
residual average point-point distance is 140 mm.

The laser data for ``Br\`{a}'' covers a larger part of the Piazza Br\`{a} site,
including the interior of the Arena, and comes as a point cloud with an average
resolution of 20 mm.  Unfortunately the laser survey contains Christmas
decoration (including a huge comet star rising from inside the Arena) and market
stalls that were not present during the photographic surveys. A detailed
comparison would entail the manual trimming of all the inconsistencies, which
would be too cumbersome.  This fact, together with a larger GSD might be the
cause of the higher residual distance obtained in this case (360 mm).

\begin{table}[tb]
\centering
\caption{Comparison with laser data. Mean and standard deviation of
  closest-points distance after registration. All measures are in mm.}
\smallskip
\begin{tabular}{lllll}
\hline & Mean  & Std Dev  & GSD\\ 
Br\`a & 360 & 300 & $\approx$ 10\\ 
S. Giacomo & 140 & 220 & $\approx$  ~4 \\ 
Duomo di Pisa & 73 & 185 &  $\approx$  ~2.8 \\ 
\hline
\end{tabular}
\label{tab:comparisonLaser}
\end{table}

\subsection{Running Times.}

The overall running times for both pipelines are reported in Table
\ref{tab:runningTimes}.  All the experiments were carried out on a workstation
equipped with a Intel Xeon W3565 cpu @ 3.20 Ghz, 36 Gb of RAM and a Nvidia
Geforce 640 GT video card.

 It should be noted that, in general, the running times are comparable. \vsfm is
 faster when the dataset is composed of a small amount of images, while \sammy
 seems to scale better when the number of image increases. This is mostly due to
 the matching phase, where by default \vsfm consider all the possible pairwise
 matches.  Please note also that the running time does not depend only on the
 number of images and the image resolution, but also on unpredictable factors
 such as the amount of overlap among the images or their content. For example,
 ``Ventimiglia'' is composed of many overlapping and high resolution images and
 the higher running time of \sammy can be probably lead back to this.

\begin{table}[htbp]
\centering
\caption{Running times [min] for \sammy and \vsfm.}
\smallskip 
\begin{tabular}{r r r r r}
  \hline           &\sammy    &\vsfm   \\
   Br\`a    & 136    & 327          \\ 
  Duomo di Pisa  & 149    & 316  \\   
   S.~Giacomo      & 103    & 122  \\  
  Navona           & 29    & 19  \\  
  Ventimiglia    & 681    & 657    \\   
  Termica        & 6    & 3   \\            
  Herz-Jesu-P25   & 5    & 3 \\    
\hline 
\end{tabular}
\label{tab:runningTimes}
\end{table}

\section{Conclusions \label{sec:conc}}

In this paper we have described several improvements to the current state of the
art in the context of uncalibrated \sam from images. Our
proposal was a hierarchical framework for \sam (\sammy), which
was demonstrated to be an improvement  over the sequential approach both in computational
complexity and with respect to the overall error containment. \sammy
constitutes the first truly scalable approach to the problem of modeling
from images, showing an almost linear complexity in the number of tie-points and
images.  

Moreover, we described a novel self-calibration approach, which coupled with our
hierarchical pipeline (\sammy) constitutes the first published example
of uncalibrated \sam for generic datasets not using external,
ancillary information. The robustness of our approach has been demonstrated on
3D model datasets both qualitatively and quantitatively.

This technology has now been transferred to a company (3Dflow srl) which
produced an industry grade implementation of \sammy that can be freely
downloaded\footnote{http://samantha.3dflow.net}.

\appendix

\section{Edge-connectedness of $G'$.}

\label{sec:app2}
Thesis: the subgraph $G'$ produced by the method reported in
Sec.~\ref{sec:broad} is $m$-edge-connected, provided that $m$ independent
spanning tree can be extracted from $G$.

To prove the thesis we rely upon the following observation.  Consider an
undirected graph $G$ with the capacity of all edges set to one; $G$ is
$k$-edge-connected if and only if the maximum flow from $u$ to $v$ is at least
$k$ for any node pair $(u,v)$.

Since our $G'$ is the union of $m$ independent (disjoint) 1-edge-connected
graphs, each of them adds an independent path with unit capacity from every node
pair $(u,v)$, so the maximum flow from every pair $(u,v)$ in $G'$ is $m$.

\subsection*{Acknowledgements}

This research project has been partially supported by a grant from the Veneto
Region (call POR CRO 2007-2013 “Regional competitiveness and employment” Action
5.1.1 “Interregional Cooperation”).

The ``Ventimiglia'' images (courtesy of Fabio Remondino) were acquired by FBK
Trento (3dom.fbk.eu) within a project funded by ARCUS SPA and supported by
Direzione Regionale per i Beni Culurali e Paesaggistici della Ligura (MiBAC) and
Soprintendenza per i Beni Archeologici della Liguria.  The laser scanning of ``
Br\`a'' was  conducted by Gexcel s.r.l. with the EU JRC - Ispra and the
permission of the municipality of Verona. The laser data of the ``Duomo di
Pisa'' comes from the ``Cattedrale Digitale'' project, while the photo set is
courtesy of the Visual Computing Lab (ISTI-CNR, Pisa). Domenico
 Visintini (Universit\`a di Udine) took the laser survey of `` S. Giacomo''.
 The images of the ``Termica'' dataset have been captured by Airmap (curtesy of
 Davide Zorzetto). Alberto Beinat and Marco Fassetta (Universit\`a di Udine)
 surveyed the GCPs.  The Navona dataset was acquired by L. Barazzetti
 (Politecnico di Milano, Italy) and used for the evaluation of automated image
 orientation methods during the ISPRS 3D-ARCH workshop in 2011.  The
 ``Herz-Jesu-P25'' images come from EPFL dense multi-view stereo evaluation
 dataset.


\begin{thebibliography}{10}
\expandafter\ifx\csname url\endcsname\relax
  \def\url#1{\texttt{#1}}\fi
\expandafter\ifx\csname urlprefix\endcsname\relax\def\urlprefix{URL }\fi
\expandafter\ifx\csname href\endcsname\relax
  \def\href#1#2{#2} \def\path#1{#1}\fi

\bibitem{ThoBroWei04}
T.~Thorm{\"a}hlen, H.~Broszio, A.~Weissenfeld, Keyframe selection for camera
  motion and structure estimation from multiple views, in: Proceedings of the
  European Conference on Computer Vision, 2004, pp. 523--535.

\bibitem{CorEA08}
N.~Cornelis, B.~Leibe, K.~Cornelis, L.~V. Gool, {3D} urban scene modeling
  integrating recognition and reconstruction, International Journal of Computer
  Vision 78~(2-3) (2008) 121--141.

\bibitem{FitZis98}
A.~W. Fitzgibbon, A.~Zisserman, Automatic camera recovery for closed and open
  image sequencese, in: Proceedings of the European Conference on Computer
  Vision, 1998, pp. 311--326.

\bibitem{SteEssDel03}
D.~Steedly, I.~Essa, F.~Dellaert, Spectral partitioning for structure from
  motion, in: Proceedings of the International Conference on Computer Vision,
  2003, pp. 649--663.

\bibitem{NiSteDel07}
K.~Ni, D.~Steedly, F.~Dellaert, Out-of-core bundle adjustment for large-scale
  {3D} reconstruction, in: Proceedings of the International Conference on
  Computer Vision, 2007, pp. 1--8.

\bibitem{Nis00}
D.~Nist\'{e}r, Reconstruction from uncalibrated sequences with a hierarchy of
  trifocal tensors, in: Proceedings of the European Conference on Computer
  Vision, 2000, pp. 649--663.

\bibitem{ShuKeZha99}
H.-Y. Shum, Q.~Ke, Z.~Zhang, Efficient bundle adjustment with virtual key
  frames: A hierarchical approach to multi-frame structure from motion, in:
  Proceedings of the IEEE Conference on Computer Vision and Pattern
  Recognition, 1999, pp. 2538--2543.

\bibitem{GibEA02}
S.~Gibson, J.~Cook, T.~Howard, R.~Hubbold, D.~Oram, Accurate camera calibration
  for off-line, video-based augmented reality, Mixed and Augmented Reality,
  IEEE / ACM International Symposium on.

\bibitem{SnaSeiSze08}
N.~Snavely, S.~M. Seitz, R.~Szeliski, Skeletal graphs for efficient structure
  from motion, in: Proceedings of the IEEE Conference on Computer Vision and
  Pattern Recognition, 2008, pp. 1--8.

\bibitem{SchZis02}
F.~Schaffalitzky, A.~Zisserman, Multi-view matching for unordered image sets,
  or "how do i organize my holiday snaps?", in: Proceedings of the 7th European
  Conference on Computer Vision, 2002, pp. 414--431.

\bibitem{Ni2012}
K.~Ni, F.~Dellaert, Hypersfm, in: Proceedings of the 2012 Second International
  Conference on 3D Imaging, Modeling, Processing, Visualization \&
  Transmission, 2012, pp. 144--151.

\bibitem{GheFarFus10}
R.~Gherardi, M.~Farenzena, A.~Fusiello, Improving the efficiency of
  hierarchical structure-and-motion, in: Proceedings of the IEEE Conference on
  Computer Vision and Pattern Recognition, San Francisco, CA, 2010, pp. 1594 --
  1600.

\bibitem{AgaEA09}
S.~Agarwal, Y.~Furukawa, N.~Snavely, I.~Simon, B.~Curless, S.~M. Seitz,
  R.~Szeliski, Building rome in a day, Communications of the ACM 54~(10) (2011)
  105--112.
\newblock \href {http://dx.doi.org/10.1145/2001269.2001293}
  {\path{doi:10.1145/2001269.2001293}}.

\bibitem{Frahm2010}
J.-M. Frahm, P.~Fite-Georgel, D.~Gallup, T.~Johnson, R.~Raguram, C.~Wu, Y.-H.
  Jen, E.~Dunn, B.~Clipp, S.~Lazebnik, M.~Pollefeys, Building {R}ome on a
  cloudless day, in: Proceedings of the 11th European conference on Computer
  vision: Part IV, 2010, pp. 368--381.

\bibitem{Arie12}
M.~Arie-Nachimson, S.~Z. Kovalsky, I.~Kemelmacher-Shlizerman, A.~Singer,
  R.~Basri, Global motion estimation from point matches, in: International
  Conference on 3D Imaging, Modeling, Processing, Visualization and
  Transmission, 2012, pp. 81--88.

\bibitem{Snavely14}
K.~Wilson, N.~Snavely, Robust global translations with 1dsfm, in: Proceedings
  of the European Conference on Computer Vision, Vol. 8691 of Lecture Notes in
  Computer Science, Springer, 2014, pp. 61--75.

\bibitem{disco11}
D.~Crandall, A.~Owens, N.~Snavely, D.~P. Huttenlocher, Discrete-continuous
  optimization for large-scale structure from motion, in: Proceedings of the
  IEEE Conf. on Computer Vision and Pattern Recognition, 2011, pp. 3001--3008.

\bibitem{Kahl08}
F.~Kahl, R.~Hartley, Multiple-view geometry under the $l_{\infty}$-norm, IEEE
  Transactions on Pattern Analysis and Machine Intelligence 30~(9) (2008)
  1603--1617.

\bibitem{MartinecP07}
D.~Martinec, T.~Pajdla, Robust rotation and translation estimation in multiview
  reconstruction, in: Proceedings of the IEEE Conference on Computer Vision and
  Pattern Recognition, 2007.

\bibitem{SinhaSS10}
S.~N. Sinha, D.~Steedly, R.~Szeliski, A multi-stage linear approach to
  structure from motion, in: Proceedings of the European Conference on Computer
  Vision, 2010, pp. 267--281.

\bibitem{Brand04}
M.~Brand, M.~Antone, S.~Teller, Spectral solution of large-scale extrinsic
  camera calibration as a graph embedding problem, in: Proceedings of the IEEE
  Conference on Computer Vision and Pattern Recognition, 2004, pp. 262--273.

\bibitem{Govindu01}
V.~M. Govindu, Combining two-view constraints for motion estimation, in:
  Proceedings of the IEEE Conference on Computer Vision and Pattern
  Recognition, 2001, pp. II--218--II--225.

\bibitem{Jiang13}
N.~Jiang, Z.~Cui, P.~Tan, A global linear method for camera pose registration,
  in: Proceedings of the International Conference on Computer Vision, 2013, pp.
  481--488.

\bibitem{Moulon13}
P.~Moulon, P.~Monasse, R.~Marlet, {Global Fusion of Relative Motions for
  Robust, Accurate and Scalable Structure from Motion}, in: Proceedings of the
  International Conference on Computer Vision, 2013, pp. 3248--3255.

\bibitem{Basri14}
O.~Ozyesil, A.~Singer, R.~Basri, \href{http://arxiv.org/abs/1312.5047}{Camera
  motion estimation by convex programming}, ArXiv e-prints 1312.5047, to appear
  in the SIAM Journal on Imaging Sciences.
\newline\urlprefix\url{http://arxiv.org/abs/1312.5047}

\bibitem{BroLow05}
M.~Brown, D.~G. Lowe, Unsupervised {3D} object recognition and reconstruction
  in unordered datasets, in: Proceedings of the International Conference on 3D
  Digital Imaging and Modeling, 2005, pp. 56--63.

\bibitem{IrsZacBis07}
A.~Irschara, C.~Zach, H.~Bischof, Towards wiki-based dense city modeling, in:
  Proceedings of the 11th International Conference on Computer Vision, 2007,
  pp. 1--8.

\bibitem{EnqvistKO11}
O.~Enqvist, F.~Kahl, C.~Olsson, Non-sequential structure from motion, in:
  Eleventh Workshop on Omnidirectional Vision, Camera Networks and
  Non-classical Camera, 2011, pp. 264--271.

\bibitem{OlssonE11a}
C.~Olsson, O.~Enqvist, Stable structure from motion for unordered image
  collections, in: Proceedings of the 17th Scandinavian conference on image
  analysis, Springer-Verlag, 2011, pp. 524--535.

\bibitem{VerGoo06}
M.~Vergauwen, L.~V. Gool, Web-based {3D} reconstruction service, Machine Vision
  and Applications 17~(6) (2006) 411--426.

\bibitem{KamEA06}
G.~Kamberov, G.~Kamberova, O.~Chum, S.~Obdrzalek, D.~Martinec, J.~Kostkova,
  T.~Pajdla, J.~Matas, R.~Sara, {3D} geometry from uncalibrated images, in:
  Proceedings of the 2nd International Symposium on Visual Computing, 2006, pp.
  802--813.

\bibitem{SnaSeiSze06}
N.~Snavely, S.~M. Seitz, R.~Szeliski, Photo tourism: exploring photo
  collections in {3D}, in: SIGGRAPH: International Conference on Computer
  Graphics and Interactive Techniques, 2006, pp. 835--846.

\bibitem{PolEA08}
M.~Pollefeys, D.~Nist\'{e}r, J.~M. Frahm, A.~Akbarzadeh, P.~Mordohai, B.~Clipp,
  C.~Engels, D.~Gallup, S.~J. Kim, P.~Merrell, C.~Salmi, S.~Sinha, S.~Sinha,
  B.~Talton, L.~Wang, Q.~Yang, H.~Stew\'{e}nius, R.~Yang, G.~Welch, H.~Towles,
  Detailed real-time urban {3D} reconstruction from video, International
  Journal of Computer Vision 78~(2-3) (2008) 143--167.

\bibitem{MayFau92}
S.~J. Maybank, O.~Faugeras, A theory of self-calibration of a moving camera,
  International Journal of Computer Vision 8~(2) (1992) 123--151.

\bibitem{Tri97}
B.~Triggs, Autocalibration and the absolute quadric, in: Proceedings of the
  IEEE Conference on Computer Vision and Pattern Recognition, Puerto Rico,
  1997, pp. 609--614.

\bibitem{PolKocGoo98}
M.~Pollefeys, R.~Koch, L.~{Van Gool}, Self-calibration and metric
  reconstruction in spite of varying and unknown internal camera parameters,
  in: Proceedings of the International Conference on Computer Vision, Bombay,
  1998, pp. 90--95.

\bibitem{PolVerVan02}
M.~Pollefeys, F.~Verbiest, L.~{Van Gool}, Surviving dominant planes in
  uncalibrated structure and motion recovery, in: Proceedings of the European
  Conference on Computer Vision, 2002, pp. 837--851.

\bibitem{HeyCip01}
Y.~Seo, A.~Heyden, R.~Cipolla, A linear iterative method for auto-calibration
  using the {DAC} equation, in: Proceedings of the IEEE Conference on Computer
  Vision and Pattern Recognition, Vol.~1, 2001, p. 880.

\bibitem{Cha07}
M.~Chandraker, S.~Agarwal, F.~Kahl, D.~Nister, D.~Kriegman, Autocalibration via
  rank-constrained estimation of the absolute quadric, in: Proceedings of the
  IEEE Conference on Computer Vision and Pattern Recognition, 2007, pp. 1--8.

\bibitem{FusBenFarBus04}
A.~Fusiello, A.~Benedetti, M.~Farenzena, A.~Busti, Globally convergent
  autocalibration using interval analysis, IEEE Transactions on Pattern
  Analysis and Machine Intelligence 26~(12) (2004) 1633--1638.

\bibitem{Bar07}
B.~Bocquillon, A.~Bartoli, P.~Gurdjos, A.~Crouzil, On constant focal length
  self-calibration from multiple views, in: Proceedings of the IEEE Conference
  on Computer Vision and Pattern Recognition, 2007.

\bibitem{Cha07b}
M.~Chandraker, S.~Agarwal, D.~Kriegman, S.~Belongie, Globally optimal affine
  and metric upgrades in stratified autocalibration, in: Proceedings of the
  International Conference on Computer Vision, 2007, pp. 1--8.

\bibitem{HarHayAgaRei99}
R.~Hartley, E.~Hayman, L.~de~Agapito, I.~Reid, Camera calibration and the
  search for infinity, in: Proceedings of the International Conference on
  Computer Vision, 1999, pp. 510--517.

\bibitem{FarFusGhe09}
M.~Farenzena, A.~Fusiello, R.~Gherardi, Structure-and-motion pipeline on a
  hierarchical cluster tree, in: IEEE International Workshop on 3-{D} Digital
  Imaging and Modeling, Kyoto, Japan, 2009, pp. 1489--1496.

\bibitem{GheFus10}
R.~Gherardi, A.~Fusiello, Practical autocalibration, in: Proceedings of the
  European Conference on Computer Vision, Lecture Notes in Computer Science,
  2010, pp. 790--801.

\bibitem{Lin98}
T.~Lindeberg, Feature detection with automatic scale selection, International
  Journal of Computer Vision 30 (1998) 79--116.

\bibitem{MS05}
K.~Mikolajczyk, C.~Schmid, A performance evaluation of local descriptors, IEEE
  Transactions on Pattern Analysis \& Machine Intelligence 27~(10) (2005)
  1615--1630.

\bibitem{MouAry07}
D.~M. Mount, S.~Arya, Ann: A library for approximate nearest neighbor
  searching, in: http://www.cs.umd.edu/~mount/ANN/, 1996.

\bibitem{Brolow03}
M.~Brown, D.~Lowe, Recognising panoramas, in: Proceedings of the 9th
  International Conference on Computer Vision, Vol.~2, 2003, pp. 1218--1225.

\bibitem{Low04}
D.~G. Lowe, Distinctive image features from scale-invariant keypoints,
  International Journal of Computer Vision 60~(2) (2004) 91--110.

\bibitem{alhwarin2010vf}
F.~Alhwarin, D.~Risti{\'c}-Durrant, A.~Gr{\"a}ser, {VF-SIFT}: very fast {SIFT}
  feature matching, in: Pattern Recognition, Springer, 2010, pp. 222--231.

\bibitem{TorZis00}
P.~H.~S. Torr, A.~Zisserman, {MLESAC}: A new robust estimator with application
  to estimating image geometry, Computer Vision and Image Understanding 78
  (2000) 2000.

\bibitem{Zha95}
Z.~Zhang, R.~Deriche, O.~Faugeras, Q.-T. Luong, A robust technique for matching
  two uncalibrated images through the recovery of the unknown epipolar
  geometry, Artificial Intelligence 78~(1-2) (1995) 87 -- 119.

\bibitem{Ste99}
C.~V. Stewart, Robust parameter estimation in computer vision, SIAM Review
  41~(3) (1999) 513--537.

\bibitem{LuoFau96}
Q.-T. Luong, O.~D. Faugeras, The fundamental matrix: Theory, algorithms, and
  stability analysis, International Journal of Computer Vision 17 (1996)
  43--75.

\bibitem{ChuPajStu05}
O.~Chum, T.~Pajdla, P.~Sturm, The geometric error for homographies, Computer
  Vision and Image Understanding 97~(1) (2005) 86--102.

\bibitem{Tor97}
P.~H.~S. Torr, An assessment of information criteria for motion model
  selection, Proceedings of the IEEE Conference on Computer Vision and Pattern
  Recognition (1997) 47--53.

\bibitem{QuaLeiGoo08}
T.~Quack, B.~Leibe, L.~Van~Gool, World-scale mining of objects and events from
  community photo collections, in: Proceedings of the International Conference
  on Content-based Image and Video Retrieval, 2008, pp. 47--56.

\bibitem{SimSnaSei07}
I.~Simon, N.~Snavely, , S.~M. Seitz, Scene summarization for online image
  collections, in: Proceedings of the Internation Conference on Computer
  Vision, 2007, pp. 1--8.

\bibitem{DudHar73}
R.~O. Duda, P.~E. Hart, Pattern Classification and Scene Analysis, John Wiley
  and Sons, 1973.

\bibitem{Har92}
R.~I. Hartley, Estimation of relative camera position for uncalibrated cameras,
  in: Proceedings of the European Conference on Computer Vision, 1992, pp.
  579--587.

\bibitem{bundle}
B.~Triggs, P.~F. McLauchlan, R.~I. Hartley, A.~W. Fitzgibbon, Bundle adjustment
  - a modern synthesis, in: Proceedings of the International Workshop on Vision
  Algorithms, 2000, pp. 298--372.

\bibitem{HarStu97}
R.~I. Hartley, P.~Sturm, Triangulation, Computer Vision and Image Understanding
  68~(2) (1997) 146--157.

\bibitem{HamRouRonSta86}
F.~Hampel, P.~Rousseeuw, E.~Ronchetti, W.~Stahel, Robust Statistics: the
  Approach Based on Influence Functions, Wiley Series in probability and
  mathematical statistics, John Wiley \& Sons, 1986.

\bibitem{GarCroFus12}
V.~Garro, F.~Crosilla, A.~Fusiello, Solving the pnp problem with anisotropic
  orthogonal procrustes analysis, in: Proceedings of the Second Joint
  3DIM/3DPVT Conference: 3D Imaging, Modeling, Processing, Visualization and
  Transmission, 2012, pp. 262--269.

\bibitem{RaguramF11}
R.~Raguram, J.-M. Frahm, Recon: Scale-adaptive robust estimation via residual
  consensus, in: Proceedings of the International Conference on Computer
  Vision, 2011, pp. 1299--1306.

\bibitem{Kan93}
K.~Kanatani, Geometric Computation for Machine Vision, Oxford University Press,
  1993.

\bibitem{ShoCar70}
P.~SchÃ¶nemann, R.~Carroll, Fitting one matrix to another under choice of a
  central dilation and a rigid motion, Psychometrika 35~(2) (1970) 245--255.

\bibitem{Kra97}
K.~Kraus, Photogrammetry: Advanced methods and applications, Vol.~2,
  D{\"u}mmler, 1997.

\bibitem{CorLeiRiv01}
T.~H. Cormen, C.~E. Leiserson, R.~L. Rivest, C.~Stein, Introduction to
  Algorithms, The MIT Press, Cambridge, MA, USA, 2001.

\bibitem{ZhaSha03}
Z.~Zhang, Y.~Shan, Incremental motion estimation through modified bundle
  adjustment, in: Proceedings of the International Conference on Image
  Processing, 2003, pp. II--343--6.

\bibitem{MouEA06}
E.~Mouragnon, M.~Lhuillier, M.~Dhome, F.~Dekeyser, P.~Sayd, Real time
  localization and 3d reconstruction, in: Proceedings of the International
  Conference on Computer Vision and Pattern Recognition, 2006, pp. 363--370.

\bibitem{LuoVie96}
Q.-T. Luong, T.~Vi\'{e}ville, Canonical representations for the geometries of
  multiple projective views, Computer Vision and Image Understanding 64~(2)
  (1996) 193--229.

\bibitem{HarZis00}
R.~Hartley, A.~Zisserman, Multiple View Geometry in Computer Vision, Cambridge
  University Press, 2003.

\bibitem{remondino2012low}
F.~Remondino, S.~Del~Pizzo, T.~P. Kersten, S.~Troisi, Low-cost and open-source
  solutions for automated image orientation--a critical overview, in: Progress
  in Cultural Heritage Preservation, Springer, 2012, pp. 40--54.

\bibitem{strecha2008}
C.~Strecha, W.~Von~Hansen, L.~Van~Gool, P.~Fua, U.~Thoennessen, On benchmarking
  camera calibration and multi-view stereo for high resolution imagery, in:
  IEEE Conference on Computer Vision and Pattern Recognition, 2008, pp. 1--8.

\bibitem{VisualSfm}
C.~Wu, Visualsfm: A visual structure from motion system,
  http://homes.cs.washington.edu/~ccwu/vsfm/.

\bibitem{MulticoreBA}
C.~Wu, S.~Agarwal, B.~Curless, S.~M. Seitz, Multicore bundle adjustment, in:
  IEEE Conference on Computer Vision and Pattern Recognition, 2011, pp.
  3057--3064.

\bibitem{CloudCompare}
Cloud compare, \url{http://www.danielgm.net/cc/}.

\end{thebibliography}

\end{document}